\pdfoutput=1

\documentclass[11pt]{article}

\usepackage[final]{acl}

\usepackage{times}
\usepackage{latexsym}

\usepackage[T1]{fontenc}
\usepackage[utf8]{inputenc}
\usepackage{algorithm}
\usepackage{algorithmic}
\usepackage{makecell}
\usepackage{amssymb}
\usepackage{hyperref}
\usepackage{color}
\definecolor{span_pink}{RGB}{255,122,179}
\definecolor{entity_blue}{RGB}{148,169,216}
\usepackage{newfloat}
\usepackage{listings}
\floatstyle{ruled}
\newfloat{listing}{tb}{lst}{}
\floatname{listing}{Listing}

\usepackage{microtype}
\usepackage{booktabs} 
\usepackage{url}
\usepackage{float}
\usepackage{graphicx}
\usepackage{graphbox} 
\usepackage{tabularx}
\usepackage{multirow}
\usepackage{multicol}
\usepackage{balance}
\usepackage{amsfonts}
\usepackage{amsmath}
\usepackage{xcolor}
\usepackage{colortbl}
\usepackage{diagbox}
\usepackage{subfigure}
\makeatletter

\newcommand{\thickhline}{%
    \noalign {\ifnum 0=`}\fi \hrule height 1pt
    \futurelet \reserved@a \@xhline
}
\newcommand*\circled[1]{\kern-2.5em%
  \put(0,4){\color{white}\circle*{18}}\put(0,4){\circle{10}}%
  \put(-3,0){\color{black}\bfseries#1}~~}
\usepackage{tikz} 

\usepackage{enumitem}

\newcommand{\printfnsymbol}[1]{%
  \textsuperscript{\@fnsymbol{#1}}%
}
\title{\includegraphics[scale=0.063, align=c]{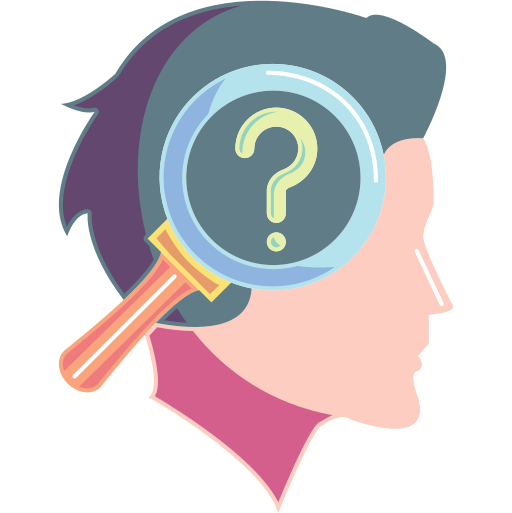} Vision-Language Introspection: Mitigating Overconfident Hallucinations in MLLMs via Interpretable Bi-Causal Steering}

\author{
    Shuliang Liu\textsuperscript{\rm 1,2},
    Songbo Yang\textsuperscript{\rm 1},
    Dong Fang\textsuperscript{\rm 3,*},
    Sihang Jia\textsuperscript{\rm 1,2},
    Yuqi Tang\textsuperscript{\rm 1}, \\
    \textbf{Lingfeng Su}\textsuperscript{\rm 3},
    \textbf{Ruoshui Peng}\textsuperscript{\rm 1,2},
    \textbf{Yibo Yan}\textsuperscript{\rm 1,2},
    \textbf{Xin Zou}\textsuperscript{\rm 1,2},
    \textbf{Xuming Hu}\textsuperscript{\rm 1,2,*} \\
    \textsuperscript{\rm 1} {The Hong Kong University of Science and Technology (Guangzhou)} \\
    \textsuperscript{\rm 2} {The Hong Kong University of Science and Technology}, \textsuperscript{\rm 3} {LIGHTSPEED} \\
    \texttt{\href{mailto:shulianglyo@gmail.com}{shulianglyo@gmail.com}},
    \texttt{\href{mailto:df572@outlook.com}{df572@outlook.com}}, \texttt{\href{mailto:xuminghu@hkust-gz.edu.cn}{xuminghu@hkust-gz.edu.cn}}
}

\begin{document}
\maketitle

\begingroup
\renewcommand\thefootnote{*}
\footnotetext{Corresponding authors.}
\endgroup

\begin{abstract}


Object hallucination critically undermines the reliability of Multimodal Large Language Models, often stemming from a fundamental failure in cognitive introspection—where models blindly trust linguistic priors over specific visual evidence. Existing mitigations remain limited: contrastive decoding approaches operate superficially without rectifying internal semantic misalignments, while current latent steering methods rely on static vectors that lack instance-specific precision. We introduce \textbf{Vision-Language Introspection (VLI)}, a training-free inference framework that simulates a metacognitive self-correction process. VLI first performs \textit{Attributive Introspection} to diagnose hallucination risks via probabilistic conflict detection and localize the causal visual anchors. It then employs \textit{Interpretable Bi-Causal Steering} to actively modulate the inference process, dynamically isolating visual evidence from background noise while neutralizing blind confidence through adaptive calibration. VLI achieves state-of-the-art performance on advanced models, reducing object hallucination rates by 12.67\% on MMHal-Bench and improving accuracy by 5.8\% on POPE.

\end{abstract}

\section{Introduction}

Multimodal Large Language Models (MLLMs) have advanced significantly in reasoning but suffer critically from object hallucination, generating plausible yet non-existent objects \citep{liu2024phd}. Recent studies identify this not merely as perceptual error, but a failure of \textit{cognitive introspection}: models exhibit blind confidence, over-relying on linguistic priors rather than verifying generation against specific visual evidence \citep{min2024mitigating, zhou2023analyzing}.

Current mitigation strategies have shifted from costly retraining \citep{ding2025pami, jiang2024hallucination, xing2024mitigating, hei2025unlocking} toward lightweight training-free paradigms \citep{chen2025decoupling, favero2024multi, wu2025generate, yin2024woodpecker, zhang2025self}, broadly bifurcating into distribution-level and representation-level interventions. However, both paradigms face fundamental limitations in \textit{precision} and \textit{cognitive depth}. Distribution-level methods, like Contrastive Decoding ~\cite{leng2024mitigating}, often indiscriminately mask visual inputs \citep{an2025mitigating, chen2024halc}, inadvertently discarding background context essential for reasoning~\citep{fu2025mitigating, zhao2025mitigating} and failing to rectify deep-seated erroneous visual-semantic connections \citep{liu2024paying, he2025cracking}. Conversely, representation-level strategies relying on static steering vectors \citep{shi2025exposing, li2025hidden, suo2025octopus} lack instance-specific granularity \citep{liu2025reducing, su2025activation, yang2025nullu}. Crucially, they fail to address the model's intrinsic \textit{blind confidence} \citep{duan2025truthprint, ye2025claim}, as generic interventions cannot decouple true visual understanding from stubborn hallucinatory biases \citep{zhu2025mitigating, kalai2025language, ling2025wakenllm}.

Crucially, existing methods treat multimodal reasoning as a binary selection between linguistic priors and visual features, overlooking the cognitive necessity of \textbf{explicit causal dependency}. Analogous to human perception, where expectations are actively verified against specific visual regions, true reasoning demands dynamic self-verification rather than static probability ranking. Current LVLMs lack this capability, causing blind confidence. While mechanistic interpretability diagnoses these states \citep{jiang2025devils}, we argue that diagnosis must be operationalized into active control to bridge the gap between passive interpretation and rectification \citep{park2025halloc, chen2025seeing, bae2025mash}.

Building on these insights, we propose \textbf{Vision-Language Introspection (VLI)}, a training-free framework simulating metacognitive self-correction. Unlike post-hoc methods \citep{chen2025rrhf, heiman2025factchexcker}, VLI employs a bidirectional mechanism aligning visual evidence with textual generation. First, \textit{Attributive Introspection} uses attention purification to isolate the causal visual anchor, strictly differentiating object pixels from background context \citep{zhao2024mitigating}. Second, \textit{Interpretable Bi-Causal Steering} rectifies inference by constructing \textbf{Anchor-Only} and \textbf{Context-Only} counterfactual states via inpainting. This derives a dynamic correction vector that enhances focus on visual evidence while suppressing background noise triggering linguistic priors. Finally, \textit{Adaptive Confidence Calibration} \citep{xie2024calibrating} addresses blind confidence by measuring cognitive conflict between holistic and counterfactual states, adaptively penalizing ungrounded certainty.

Extensive experiments demonstrate that VLI significantly outperforms the baseline by reducing MMHal hallucination rates by up to 12.67\% and enhancing POPE accuracy by 6.33\%, while surpassing state-of-the-art methods by margins of 5.37\% and 1.60\%, respectively.
Our contributions are:
\begin{itemize}
    \item We propose \textbf{VLI}, a framework that systematically diagnoses and rectifies object hallucinations by interpreting and manipulating specific visual anchors.
    \item We introduce \textbf{Bi-Causal Steering}, which performs precise latent interventions by dynamically contrasting Anchor-Only vs. Context-Only representations to isolate and reinforce the true visual cause.
    \item We integrate \textbf{Adaptive Confidence Calibration} to detect cognitive conflict during inference, preventing hallucinations driven by blind confidence.
\end{itemize}

\section{Related Work}
\label{sec:related_work}

Hallucination mitigation strategies for Multimodal Large Language Models generally categorize into training-based alignment and training-free inference intervention. We focus on the training-free paradigm to avoid prohibitive costs \citep{ding2025pami,fu2025mitigating,liu2025survey}. These approaches typically involve surface-level decoding manipulation \citep{jiang2024hallucination,ren2025knowrl,hu2022hiure,huo2025pmark,zhang2025cohemark,liu2025vla} or deep latent state intervention \citep{wu2025antidote,wu2025generate,xiao2025detecting, huang2025video}.

\subsection{Inference-Time Hallucination Mitigation}
\label{sec:rel_mitigation}

\paragraph{Decoding Strategies.}
This stream rectifies hallucinations by externally calibrating output probabilities. \textit{Contrastive Decoding} mitigates linguistic priors by contrasting original logits against distorted ones (VCD \citep{leng2024mitigating}) or utilizing decoupled projectors (IBD \citep{zhu2025ibd}, DCD \citep{chen2025decoupling}). Recent works like CICD \citep{zhao2025cross} and DeGF \citep{zhang2025self} employ cross-image references or generative feedback \citep{an2025mitigating,lee2025retrieval} to preserve visual details \citep{lyu2024alleviating,shi2025exposing}.
To improve flexibility, \textit{Adaptive Strategies} such as Octopus \citep{suo2025octopus} and MoD \citep{chen2025mixture} dynamically route decoding strategies, while DLC \citep{chen2025mitigating} and M3ID \citep{favero2024multi} perform real-time logit calibration.
Additionally, \textit{Penalty-Based Mechanisms} like OPERA \citep{huang2024opera} and DOPRA \citep{wei2024dopra} modify beam search to penalize over-trust patterns \citep{li2025visual}. However, these decoding methods function primarily as surface-level regularizers, failing to rectify the corrupted internal representations \citep{kaul2024throne}.

\paragraph{Latent Space Steering.}
A more intrinsic paradigm directly modulates hidden states. Methods such as VTI \citep{liu2025reducing} and Nullu \citep{yang2025nullu} employ global steering vectors derived from feature averaging or null-space projection. Others like VaLSe \citep{chen2025seeing} and TruthPrInt \citep{duan2025truthprint} use probes for guided intervention \citep{park2025halloc, zhang2025bert}. Unlike these approaches that rely on static, dataset-level vectors, our VLI framework introduces \textit{Interpretable Bi-Causal Steering}. We compute a precise, dynamic steering vector derived from the cognitive gap between counterfactual states, enabling the model to actively introspect and correct instance-specific visual grounding.

\subsection{Mechanistic Interpretability for Visual Grounding}
\label{sec:rel_interpretability}

Our work is grounded in mechanistic interpretability, which diagnoses internal attention allocation. Studies reveal that LVLMs rely on specific expert heads for semantic tracking \citep{zhao2025aligning,deng2025maskcd} but often suffer from visual attention sinks \citep{kang2025see}. While existing methods utilize these findings for passive analysis or re-weighting \citep{liu2024paying,jiang2025devils,tang2025seeing}, VLI operationalizes them into active control. 

\begin{figure*}[h]
    \centering
    \includegraphics[width=1\linewidth]{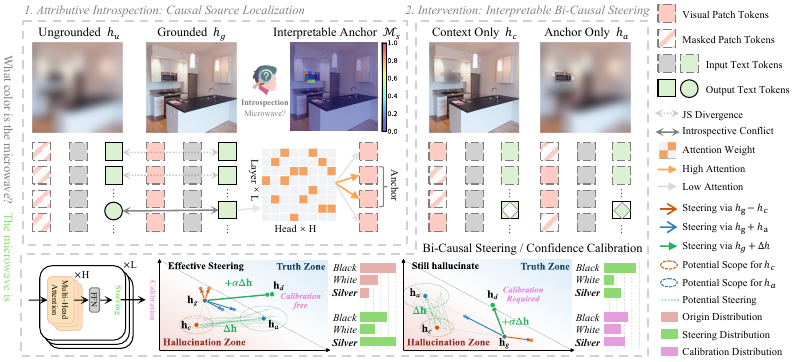}
    \caption{
    Overview of \textbf{VLI} framework.  VLI first detects \textit{Introspective Conflict} (\S\ref{sec:suspect_token}) between grounded ($h_g$) and ungrounded ($h_u$) paths to localize the causal anchor $\mathcal{M}_s$ (\S\ref{sec:anchor_extraction}) via purified expert attention (\S\ref{sec:attention_purification}, Fig.~\ref{fig:heatmap}). 
It then applies \textit{Bi-Causal Steering} (\S\ref{sec:intervention}) using the robust difference vector ($h_a - h_c$), which counters the scope instability of individual counterfactual states (Fig.~\ref{fig:Layer-wise analysis of hidden state shifts}). 
Finally, \textit{Adaptive Confidence Calibration} (\S\ref{sec:calibration}) penalizes blind confidence to mitigate persistent hallucinations.
    }
    \label{fig:overview}
    \vspace{-0.1in}
\end{figure*}

\section{Methodology}
\label{sec:method}

We introduce \textbf{Vision-Language Introspection (VLI)}, a training-free inference-time framework designed to mitigate overconfident hallucinations, illustrated in Fig.~\ref{fig:overview}. Unlike standard decoding interventions that passively suppress likely tokens, VLI simulates a metacognitive self-verification process. It addresses the fundamental disconnect between linguistic priors and visual evidence through a bidirectional mechanism: 1) \textbf{Attributive Introspection}, a diagnostic phase that traces high-risk predictions back to their causal visual origins; and 2) \textbf{Bi-Causal Steering}, an intervention phase that dynamically isolates the specific visual evidence from background noise to rectify latent representations across all model layers and calibrate blind confidence. Both process is analized with Case Study in Appendix \ref{app:case_study}.

\subsection{Attributive Introspection: Causal Source Localization}
\label{sec:suspect}

During the generation of the $t$-th token, the objective of this phase is to introspect the model's reasoning process. We aim to trace the cognitive dissonance between the model's internal priors and the actual visual input back to specific image regions, formalizing this causal origin as a pixel-precise source anchor mask $\mathcal{M}_s$.

\subsubsection{Introspective Conflict Detection}
\label{sec:suspect_token}

To quantify hallucination risks, we evaluate the consistency between visual evidence and linguistic priors via a comparative analysis of internal model states. We define two parallel decoding paths at time step $t$, tracking hidden states across all $L$ layers.

First, the \textbf{Grounded Path} represents the standard reasoning process where the model receives complete visual features $V = \mathcal{E}_V(I)$ and textual context $T_{<t}$. For each layer $l \in \{1, \dots, L\}$:
\begin{equation}
    h_{g, l}^{(t)} = \mathcal{F}_{\text{VLM}}^{(l)}(V \oplus \mathcal{E}_T(T_{<t}), h_{g, l-1}^{(t)})
    \label{eq:grounded_path}
\end{equation}
where $\mathcal{F}_{\text{VLM}}^{(l)}$ denotes the operation of the $l$-th transformer layer, and $h_{g, L}^{(t)}$ is the final latent state. The probability distribution $P_g^{(t)}$ over the vocabulary $\mathcal{V}$ is obtained via a linear projection $W_o$ followed by a Softmax operation $\sigma$:
\begin{equation}
    P_g^{(t)} =  \sigma(W_o h_{g, L}^{(t)})
\end{equation}

Second, the \textbf{Ungrounded Path} simulates reasoning relying solely on linguistic priors by masking the visual input (replacing $V$ with null tokens $\emptyset$), yielding hidden states $h_{u, l}^{(t)}$ and distribution $P_u^{(t)}$:
\begin{equation}
    P_u^{(t)} =  \sigma(W_o  \mathcal{F}_{\text{VLM}}^{(l)}(\emptyset \oplus \mathcal{E}_T(T_{<t}), h_{u, l-1}^{(t)}))
    \label{eq:l_u}
\end{equation}

We posit that the divergence between these two paths, termed \textbf{Introspective Conflict}, serves as a robust proxy for hallucination. We calculate the hallucination risk score $\mathcal{C}$ using the Jensen-Shannon (JS) divergence:
\begin{equation}
\begin{aligned}
    \mathcal{C}(P_g^{(t)}, P_u^{(t)}) = H\left(\frac{P_g^{(t)} + P_u^{(t)}}{2}\right) - \\ \frac{1}{2} [H(P_g^{(t)}) + H(P_u^{(t)})]
\end{aligned}
\label{eq:jsd}
\end{equation}
where $H(P)$ is the Shannon entropy. Upon detecting high risk (i.e., $\mathcal{C}$ exceeds a predefined threshold $\theta$, which is analyzed in Sec \ref{sec:hyperparam}) and Appendix \ref{app:introspect_case_study}, we identify the most suspicious token $t_s$. This token is defined as the vocabulary item exhibiting the maximal logarithmic divergence between the grounded and ungrounded probabilities:
\begin{equation}
t_s = \underset{w \in \mathcal{V}}{\text{argmax}} (\log P_g^{(t)}(w) - \log P_u^{(t)}(w))
\label{eq:t_s}
\end{equation}

\subsubsection{Causal Attention Purification}
\label{sec:attention_purification}

Having identified the locus of conflict $t_s$, we aim to eliminate systemic biases to extract the authentic visual evidence supporting $t_s$. We focus on identifying reliable attention heads to filter noise.
Inspired by SEVI~\cite{zhao2025aligning}, we posit that only a subset of expert heads maintain reliable causal links between semantics and visual regions. We perform offline calibration on a validation set $\mathcal{D}_{val}$. Let $\mathbf{A}_{l,h}(t_s, V) \in \mathbb{R}^{N_v}$ be the attention distribution of the $h$-th head in the $l$-th layer over visual tokens $V$ when generating $t_s$. We define the \textit{localization accuracy score} $\mu_{l,h}$ as the expected probability mass falling within the ground-truth region $R_{gt}(t_s)$:
\begin{equation}
    \mu_{l,h} = \mathbb{E}_{(I, t_s) \sim \mathcal{D}_{val}} \left[ \sum_{j=1}^{N_v} \mathbf{A}_{l,h}(t_s)_j \cdot \mathbb{I}(v_j \in R_{gt}(t_s)) \right]
\end{equation}
We select the top $M$ heads maximizing $\mu_{l,h}$ to construct the expert set $\mathcal{H}_{expert}$, following ~\cite{zhao2025aligning}. To obtain the purified attention map, we aggregate the attention weights solely from these expert heads, effectively suppressing noise and attention sinks (see Appendix \ref{app:sink} for robustness analysis against visual sinks). The unnormalized purified heatmap $\tilde{H}_{att}$ is calculated as:
\begin{equation}
    \tilde{H}_{att}(t_s) = \sum_{(l,h) \in \mathcal{H}_{expert}} \mathbf{A}_{l,h}(t_s, V)
\end{equation}
The final normalized attention distribution is $H_{att}(t_s) = \frac{\tilde{H}_{att}(t_s)}{\sum_{j=1}^{N_v} \tilde{H}_{att}(t_s)_j}$.

\subsubsection{Interpretable Anchor Extraction}
\label{sec:anchor_extraction}

To accommodate the diversity of attention distributions, we employ a \textbf{Cumulative Energy Thresholding} strategy controlled by a single hyperparameter $\rho$ (set to $0.4$, which is analyzed in Sec \ref{sec:hyperparam}). Let $\mathbf{h}_{sorted}$ denote the flattened and descendingly sorted vector of $H_{att}(t_s)$. We identify the minimal set of top-ranking pixels required to capture a total energy proportion of $\rho$:
\begin{equation}
    k = \underset{i}{\text{argmin}} \sum_{j=1}^{i} \mathbf{h}_{sorted}[j] \ge \rho \cdot \sum_{n=1}^{N_v} \mathbf{h}_{sorted}[n]
\end{equation}
The final binary causal mask is generated by selecting these top-$k$ pixels: $\mathcal{M}_s = \mathbb{I}(H_{att}(t_s) \ge \mathbf{h}_{sorted}[k])$. This method adaptively locks onto the visual regions constituting the primary semantics based on energy concentration, effectively filtering long-tail noise without requiring a pixel count.

\subsection{Intervention: Interpretable Bi-Causal Steering}
\label{sec:suppress}

The previous phase successfully diagnosed the hallucination source by isolating the causal anchor $\mathcal{M}_s$. Building upon this, the Intervention phase actively corrects the model's internal representations. We introduce \textbf{Bi-Causal Steering}, which constructs counterfactual representations to steer the model's focus toward verified visual evidence across all layers.

\subsubsection{Counterfactual Causal Construction}
\label{sec:counterfactual_input}

We utilize a pre-trained inpainting model $\mathcal{I}(\cdot, \cdot)$ to construct two complementary counterfactual visual inputs. First, the \textbf{Context-Only Image ($I_c$)} retains only the context: $I_c = \mathcal{I}(I, \mathcal{M}_s)$. Second, the \textbf{Anchor-Only Image ($I_a$)} preserves only the interpretable anchor: $I_a = \mathcal{I}(I, 1 - \mathcal{M}_s)$. Their corresponding features $V_c$ and $V_a$ are extracted via the visual encoder $\mathcal{E}_V$:
\begin{equation}
V_c = \mathcal{E}_V(I_c) \quad ; \quad V_a = \mathcal{E}_V(I_a)
\label{eq:V_c_V_a}
\end{equation}

\subsubsection{Layer-wise Bi-Causal Steering}
\label{sec:intervention}

We feed $V_c$ and $V_a$ into the VLM decoder. Unlike simple post-hoc interventions, we intervene at every layer $l \in \{1, \dots, L\}$ to fundamentally rectify the reasoning trajectory. We obtain the layer-wise context-driven states $h_{c, l}^{(t)}$ and anchor-driven states $h_{a, l}^{(t)}$, where $t$ is introspection conflict step:
\begin{equation}
\begin{aligned}
h_{c, l}^{(t)} = \mathcal{F}_{\text{VLM}}^{(l)}(V_c \oplus \mathcal{E}_T(T_{<t}), h_{c, l-1}^{(t)}) \\
h_{a, l}^{(t)} = \mathcal{F}_{\text{VLM}}^{(l)}(V_a \oplus \mathcal{E}_T(T_{<t}), h_{a, l-1}^{(t)})
\label{eq:h_c_h_a}
\end{aligned}
\end{equation}
We define a \textbf{correction vector} $\Delta_{h, l}^{(t)}$ for each layer to isolate the pure semantic information contributed by the anchor $\mathcal{M}_s$:
\begin{equation}
\Delta_{h, l}^{(t)} = h_{a, l}^{(t)} - h_{c, l}^{(t)}
\label{eq:delta_h}
\end{equation}
This vector is injected into the original grounded path at every layer. The debiased state $h_{d, l}^{(t)}$ is computed as:
\begin{equation}
h_{d, l}^{(t)} = h_{g, l}^{(t)} + \alpha \cdot \Delta_{h, l}^{(t)}
\label{eq:h_debiased}
\end{equation}
This multi-layer steering reinforces the model's perception of the key visual region $\mathcal{M}_s$ throughout the entire depth of the network, suppressing biases before they propagate to the final output.

\subsubsection{Adaptive Confidence Calibration}
\label{sec:calibration}

We introduce Adaptive Confidence Calibration to mitigate stubborn hallucinations where the model exhibits blind certainty despite lacking distinct visual support. This failure mode is characterized by high global introspection conflict $\mathcal{C}(P_g^{(t)}, P_u^{(t)})$ co-occurring with negligible local divergence between anchor and context states $\mathcal{C}(P_a^{(t)}, P_c^{(t)})$, implying the prediction relies on internal priors rather than the visual anchor.

To suppress this ungrounded confidence, we compute a calibration scalar $T_{c}$ controlled by a single risk tolerance threshold $\lambda$. The penalty activates only when the relative risk exceeds $\lambda$, bounded by a hyperbolic tangent to prevent distribution collapse:
\begin{equation}
T_{c} = 1 + \tanh \left( \max\left(0, \frac{\mathcal{C}(P_g^{(t)}, P_u^{(t)})}{\mathcal{C}(P_a^{(t)}, P_c^{(t)}) + \epsilon} - \lambda \right) \right)
\label{eq:T_calib}
\end{equation}

where $\epsilon = 10^{-6}$ is a smoothing term. This mechanism adaptively flattens the distribution only when the linguistic prior dominates the visual evidence beyond the allowed tolerance $\lambda$. 
The final corrected probability distribution is obtained by scaling the debiased distribution:
\begin{equation}
P_{\text{corr}}^{(t)} =\sigma( T_{c}^{-1} \cdot W_o h_{d, L}^{(t)}),
\label{eq:h_corr}
\end{equation}
from which the final token is decoded. Theoretical analysis for the invention process is detailed in Appendix \ref{app:theory}, while latency in Appendix \ref{app:latency}.




\definecolor{mygreen}{RGB}{20, 140, 20}
\definecolor{myred}{RGB}{180, 20, 20}

\newcommand{\upgood}[1]{\textcolor{mygreen!80!gray}{\tiny $\uparrow$#1}}
\newcommand{\downbad}[1]{\textcolor{myred!80!gray}{\tiny $\downarrow$#1}}

\newcommand{\downgood}[1]{\textcolor{mygreen!80!gray}{\tiny \hspace{1pt}$\downarrow$#1}}
\newcommand{\upbad}[1]{\textcolor{myred!80!gray}{\tiny \hspace{1pt}$\uparrow$#1}}

\newcommand{\nochange}[1]{\textcolor{gray!70}{\tiny $\sim$#1}}

\begin{table*}[htbp]
\centering
\resizebox{\textwidth}{!}{
\begin{tabular}{c c c c c c c c c c}
\toprule
\multirow{2}{*}{Model} & \multirow{2}{*}{Method} & \multicolumn{2}{c}{\textbf{MMHAL}} & \multicolumn{2}{c}{\textbf{POPE (MSCOCO)}} & \multicolumn{2}{c}{\textbf{POPE (A-OKVQA)}} & \multicolumn{2}{c}{\textbf{POPE (GQA)}} \\
\cmidrule(lr){3-4} \cmidrule(lr){5-6} \cmidrule(lr){7-8} \cmidrule(lr){9-10}
 & & Hallu. Rate $\downarrow$ & Score $\uparrow$ & Acc (\%) $\uparrow$ & F1 (\%) $\uparrow$ & Acc (\%) $\uparrow$ & F1 (\%) $\uparrow$ & Acc (\%) $\uparrow$ & F1 (\%) $\uparrow$ \\
\midrule
\rowcolor{gray!15}
\multirow{8}{*}{LLaVA-1.5} 
 & Origin & 58.30 \nochange{0.0} & 2.33 \nochange{0.00} & 83.82 \nochange{0.00} & 84.18 \nochange{0.00} & 79.54 \nochange{0.00} & 79.81 \nochange{0.00} & 77.26 \nochange{0.00} & 77.58 \nochange{0.00} \\
 & VCD & 63.54 \upbad{5.24} & 2.46 \upgood{0.13} & 84.67 \upgood{0.85} & 85.14 \upgood{0.96} & 80.43 \upgood{0.89} & 80.92 \upgood{1.11} & 78.13 \upgood{0.87} & 78.54 \upgood{0.96} \\
 & CICD & 58.33 \upbad{0.03} & 2.19 \downbad{0.14} & 86.46 \upgood{2.64} & 87.13 \upgood{2.95} & 82.14 \upgood{2.60} & 82.93 \upgood{3.12} & 79.54 \upgood{2.28} & 80.13 \upgood{2.55} \\
 & ClearSight & 57.29 \downgood{1.01} & 2.16 \downbad{0.17} & 88.74 \upgood{4.92} & 88.41 \upgood{4.23} & 84.58 \upgood{5.04} & 84.23 \upgood{4.42} & 81.89 \upgood{4.63} & 81.64 \upgood{4.06} \\
 & OPERA & 58.30 \nochange{0.0} & 2.40 \upgood{0.07} & 88.34 \upgood{4.52} & 87.96 \upgood{3.78} & 84.13 \upgood{4.59} & 83.77 \upgood{3.96} & 81.33 \upgood{4.07} & 80.97 \upgood{3.39} \\
 & VTI & 51.00 \downgood{7.30} & 2.39 \upgood{0.06} & 87.95 \upgood{4.13} & 87.69 \upgood{3.51} & 83.87 \upgood{4.33} & 83.54 \upgood{3.73} & 81.14 \upgood{3.88} & 80.88 \upgood{3.30} \\
 & Nullu & 54.17 \downgood{4.13} & 2.30 \downbad{0.03} & 87.18 \upgood{3.36} & 86.84 \upgood{2.66} & 83.02 \upgood{3.48} & 82.54 \upgood{2.73} & 80.43 \upgood{3.17} & 79.92 \upgood{2.34} \\
\rowcolor{gray!15}
 & VLI (Ours) & \textbf{45.63} \downgood{12.67} & \textbf{3.11} \upgood{0.78} & \textbf{89.61} \upgood{5.79} & \textbf{89.27} \upgood{5.09} & \textbf{85.87} \upgood{6.33} & \textbf{85.54} \upgood{5.73} & \textbf{83.49} \upgood{6.23} & \textbf{83.18} \upgood{5.60} \\
\midrule
 \rowcolor{gray!15}
\multirow{8}{*}{Qwen3-VL} 
 & Origin & 40.63 \nochange{0.0} & 3.56 \nochange{0.00} & 91.14 \nochange{0.00} & 90.53 \nochange{0.00} & 87.13 \nochange{0.00} & 86.82 \nochange{0.00} & 82.34 \nochange{0.00} & 81.93 \nochange{0.00} \\
 & VCD & 37.50 \downgood{3.13} & 3.80 \upgood{0.24} & 91.92 \upgood{0.78} & 91.37 \upgood{0.84} & 87.68 \upgood{0.55} & 87.33 \upgood{0.51} & 84.84 \upgood{2.50} & 84.48 \upgood{2.55} \\
 & CICD & 36.46 \downgood{4.17} & 3.76 \upgood{0.20} & 91.43 \upgood{0.29} & 91.08 \upgood{0.55} & 87.53 \upgood{0.40} & 87.18 \upgood{0.36} & 84.13 \upgood{1.79} & 83.78 \upgood{1.85} \\
 & ClearSight & 39.58 \downgood{1.05} & 3.55 \downbad{0.01} & 85.04 \downbad{6.10} & 83.18 \downbad{7.35} & 81.44 \downbad{5.69} & 79.82 \downbad{7.00} & 79.24 \downbad{3.10} & 78.08 \downbad{3.85} \\
 & OPERA & 39.10 \downgood{1.53} & 3.72 \upgood{0.16} & 90.87 \downbad{0.27} & 90.28 \downbad{0.25} & 86.93 \downbad{0.20} & 86.38 \downbad{0.44} & 83.54 \upgood{1.20} & 83.13 \upgood{1.20} \\
 & VTI & 36.46 \downgood{4.17} & 3.68 \upgood{0.12} & 90.62 \downbad{0.52} & 89.94 \downbad{0.59} & 86.53 \downbad{0.60} & 85.83 \downbad{0.99} & 82.68 \upgood{0.34} & 81.97 \upgood{0.04} \\
 & Nullu & 39.58 \downgood{1.05} & 3.53 \downbad{0.03} & 88.76 \downbad{2.38} & 87.62 \downbad{2.91} & 84.93 \downbad{2.20} & 83.53 \downbad{3.29} & 81.48 \downbad{0.86} & 80.93 \downbad{1.00} \\
\rowcolor{gray!15}
 & VLI (Ours) & \textbf{34.38} \downgood{6.25} & \textbf{4.32} \upgood{0.76} & \textbf{92.58} \upgood{1.44} & \textbf{92.19} \upgood{1.66} & \textbf{89.23} \upgood{2.10} & \textbf{88.79} \upgood{1.97} & \textbf{86.47} \upgood{4.13} & \textbf{85.96} \upgood{4.03} \\
\bottomrule
\end{tabular}
}
\caption{Performance evaluation on MMHal and POPE with delta improvements compared to Origin. For POPE, we report the average Accuracy and F1-score across Random, Popular, and Adversarial settings. The best results are highlighted in \textbf{bold}.}
\label{tab:merged_results}
\end{table*}

\section{Experiments}

\subsection{Experimental Settings}

\noindent\textbf{Benchmarks and Metrics.}\quad 
To comprehensively evaluate the effectiveness of our proposed \textbf{VLI} framework, we conduct experiments on two complementary benchmarks: POPE \cite{li2023evaluating} and MMHal-Bench \cite{sun2024aligning}. These benchmarks enable a systematic assessment of the model’s cognitive introspection capabilities in both discriminative and generative settings.

For POPE, which focuses on object-level discrimination, we follow standard evaluation protocols and report Accuracy and the $F_1$ score. This benchmark primarily measures the model’s ability to correctly identify the presence or absence of objects, providing a fine-grained evaluation of hallucination in discriminative tasks.

However, real-world applications often require free-form text generation rather than binary decisions. To assess hallucination under such scenarios, we further adopt MMHal-Bench, a comprehensive benchmark specifically designed for quantifying the presence and types of hallucinations in complex, open-ended VQA tasks. In MMHal-Bench, model outputs are evaluated by GPT-4 as an automated judge through comparisons with ground-truth object annotations and human-annotated captions. The benchmark reports an overall hallucination score ranging from 0 to 6 following ~\citet{liu2025reducing}, along with a detailed categorization of different hallucination types (e.g., Attributes, Adversarial, Relations).
 
\noindent\textbf{Models and Implementation Details.}\quad 
We evaluate the proposed VLI framework on two mainstream large vision–language models, LLaVA-1.5~\cite{liu2023visualinstructiontuning} and Qwen3-VL~\cite{Qwen3-VL}, using greedy decoding as the default inference strategy. VLI is a training-free, inference-time method that requires no parameter updates. All experiments are implemented in PyTorch. Unless otherwise specified, we follow the same decoding and evaluation protocol for all compared methods. For VLI, we set the anchor energy ratio $\rho=0.4$ for Attributive Introspection, the introspection conflict threshold $\theta$ to $0.1$, and the latent steering strength $\alpha=0.5$ in all experiments, as determined on the validation set.

\noindent\textbf{Baselines.}\quad 
We further compare our approach with advanced baselines spanning three representative hallucination-mitigation paradigms: (i) contrastive decoding methods, including VCD~\cite{leng2024mitigating} and CICD~\cite{zhao2025cross}; (ii) attention-intervention methods, including ClearSight ~\cite{yin2025clearsight} and OPERA~\cite{huang2024opera}; and (iii) latent-space intervention methods, including VTI~\cite{liu2025reducing} and Nullu~\cite{yang2025nullu}. For all baselines on LLaVA-1.5, we follow the original papers’ hyperparameter settings for implementation.
\begin{figure}
    \centering
    \includegraphics[width=1\linewidth]{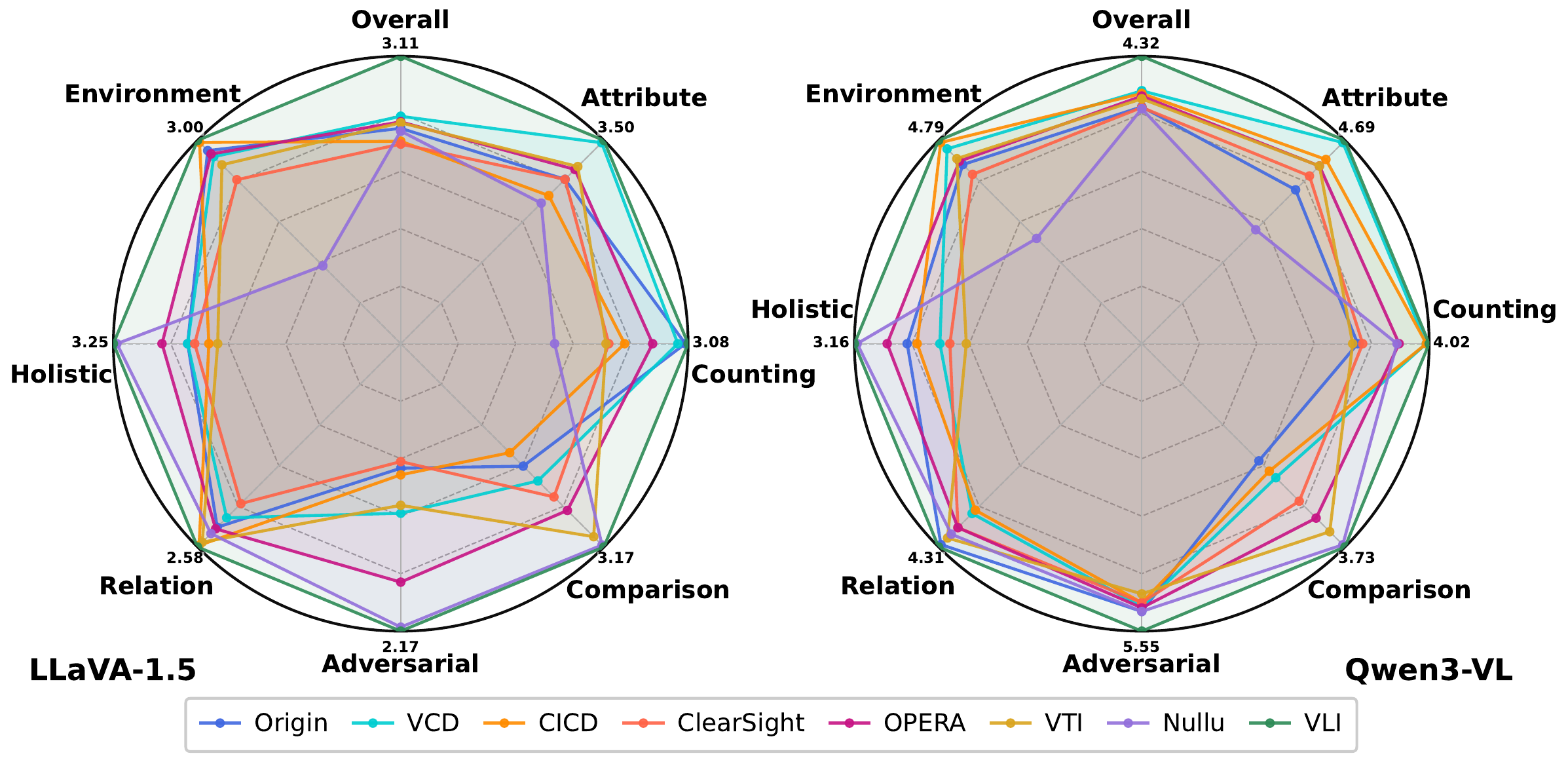}
    \caption{Detailed performance of different models on the eight categories in MMHAL-BENCH, where “Overall” indicates the averaged performance across all categories. A higher score indicates that the generated response contains fewer hallucinations and more information.}
    \label{fig:rador_chart}
\end{figure}

\subsection{Main Results}
\noindent\textbf{Performance on MMHal-Bench.}\quad
Table~\ref{tab:merged_results} demonstrates that VLI achieves state-of-the-art performance on the challenging open-ended MMHal-Bench. 
On LLaVA-1.5, VLI reduces the Hallucination Rate (HR) by a substantial \textbf{12.67\%} (from 58.30\% to 45.63\%), while on Qwen3-VL, it achieves a record low HR of \textbf{34.38\%} and the highest overall Score of \textbf{4.32}.
As illustrated in Fig.~\ref{fig:rador_chart}, VLI yields consistent gains across difficult subsets like \textit{Attribute} and \textit{Adversarial}.
These results validate the effectiveness of our \textit{Interpretable Bi-Causal Steering}: unlike decoding methods that passively penalize tokens, our mechanism actively rectifies latent visual-semantic misalignments by isolating specific visual anchors from background noise, which is critical for mitigating fine-grained hallucinations in complex open-ended generation.

\noindent\textbf{Performance on POPE.}\quad
VLI demonstrates robust generalization across all POPE datasets (MSCOCO, A-OKVQA, and GQA), outperforming baselines in discriminative tasks.
On LLaVA-1.5, VLI improves Accuracy by \textbf{5.79\%} on MSCOCO, and notably achieves even larger gains of \textbf{6.33\%} and \textbf{6.23\%} on the more challenging A-OKVQA and GQA datasets, respectively.
This superior performance on out-of-distribution and visually complex datasets highlights the advantage of our \textit{Attributive Introspection} mechanism.
By precisely localizing causal pixel evidence prior to intervention, VLI avoids the precision-recall trade-off common in global penalty-based decoding (e.g., VCD), allowing it to confidently reject non-existent objects without suppressing valid visual details.
Even on the robust Qwen3-VL, VLI further pushes accuracy to \textbf{92.58\%} on POPE-MSCOCO, proving that introspective grounding remains essential even for stronger base models.

\begin{table}[t]
\centering
\resizebox{0.98\linewidth}{!}{
\begin{tabular}{lcc}
\toprule
Method & Hallu. Rate $\downarrow$ & Score $\uparrow$ \\
\midrule
Origin & 58.30 \textcolor{gray}{\tiny \hspace{1pt}$\sim$0.0} & 2.33 \textcolor{gray}{\tiny $\sim$0.0} \\
VLI (Ours) & \textbf{45.63} \downgood{12.67} & \textbf{3.11} \upgood{0.78} \\
w/o Calibration & 47.10 \downgood{11.20} & 3.02 \upgood{0.69} \\
w/o Context only & 50.25 \downgood{8.05} & 2.82 \upgood{0.49} \\
w/o Anchor only & 53.40 \downgood{4.90} & 2.61 \upgood{0.28} \\
\bottomrule
\end{tabular}
}
\caption{Ablation study results on MMHAL-Bench with LLaVA-1.5, with delta improvements compared to Origin.}
\label{tab:ablation_study}
\end{table}

\subsection{Ablation Study}
To validate the effectiveness of our framework, we conducted an ablation study on MMHal-Bench with LLaVA-1.5 (Table~\ref{tab:ablation_study}). The full VLI yields the best performance (45.63\% hallucination rate, 3.11 score), confirming the synergy of all components.

\textbf{Dominance of Bi-Causal Steering.}
The most significant performance degradation occurs in the \textit{w/o Anchor only} setting, where the hallucination rate spikes by 7.77\%. This drop outweighs that of the \textit{w/o Context only} variant, indicating that explicitly reinforcing the visual anchor is the primary driver for error rectification. This observation aligns with the layer-wise analysis in Fig.~\ref{fig:Layer-wise analysis of hidden state shifts}, which shows that  \textit{Anchor-only} induces a much larger shift in hidden states than \textit{Context-only}. This confirms that the anchor provides the dominant semantic guidance, effectively pulling latent states away from linguistic priors toward the visual ground truth.

\textbf{Role of Calibration.}
Conversely, removing \textit{Adaptive Confidence Calibration} results in a relatively minor performance decrease (+1.47\%). This is expected, as calibration operates by smoothing the output distribution to penalize ungrounded certainty. Unlike steering, which fundamentally repairs internal semantic representations, calibration serves as a final refinement to prevent the model from being overconfident in its remaining errors.

\subsection{Visualization for Introspection and Inventation}
\paragraph{Expert Head Attention for Introspection}
\begin{figure}[h!]
    \centering
    \includegraphics[width=0.8\linewidth]{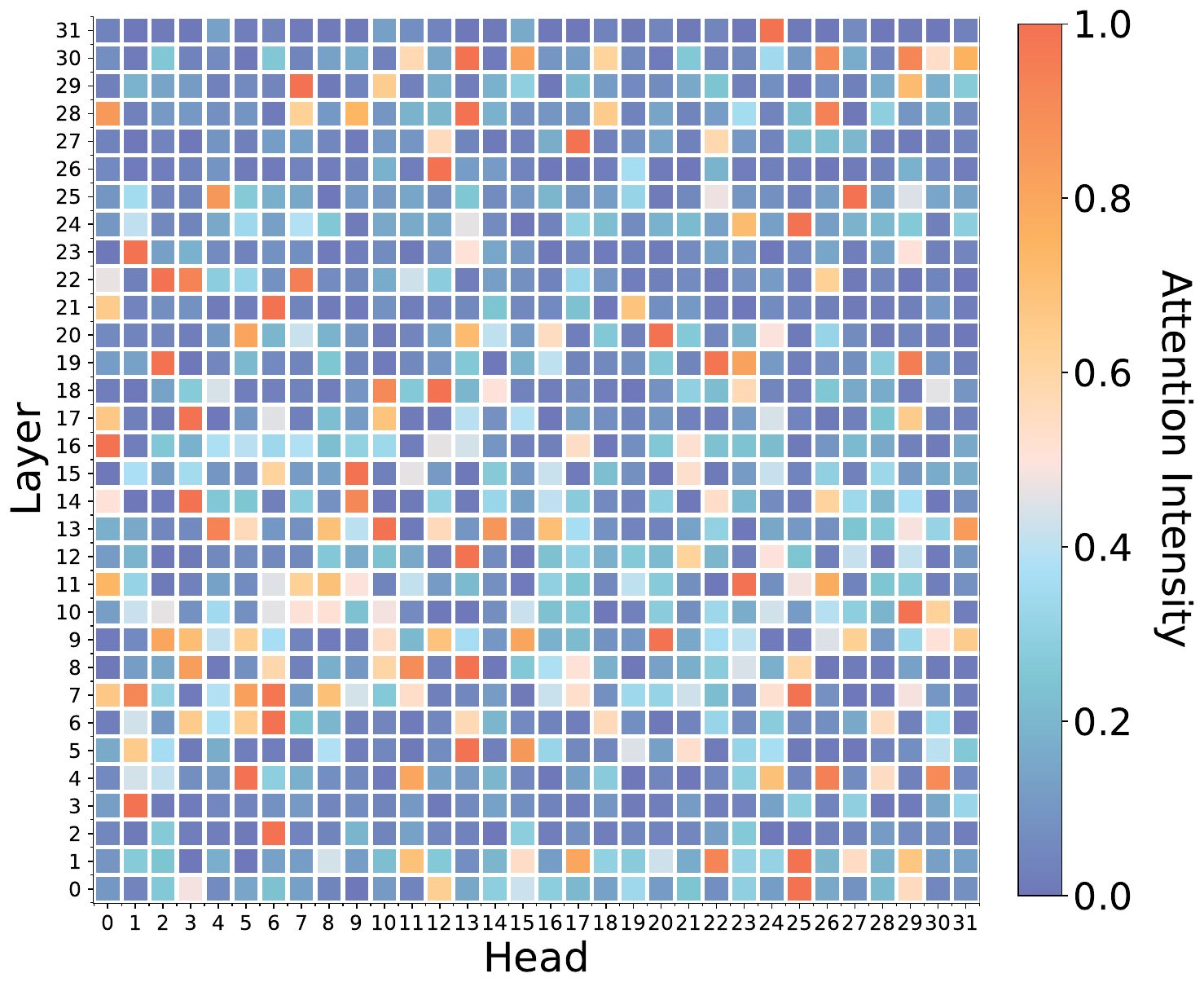}
    \caption{Heatmap of max-pooled attention intensity across layers and heads.  }
    \label{fig:heatmap}
\end{figure}
As illustrated in Fig. \ref{fig:heatmap}, the attention distribution exhibits a pronounced functional specificity and discreteness. Contrary to a uniform engagement of neural resources, we observe that the vast majority of attention heads remain distinctively silent, represented by cool colors, while high-magnitude activations are concentrated in a sparse subset of layer-specific expert heads.
This observation provides strong empirical support for our dynamic head selection mechanism. Since critical semantic information is isolated within these few expert heads, a holistic or average-based approach would inevitably introduce significant noise from the inactive majority. Consequently, dynamically identifying and prioritizing these expert heads is paramount. It allows the model to effectively filter out background interference and establish a precise causal link between the introspection conflict tokens and the relevant visual patches. By focusing solely on these high-activation pathways, our method ensures that the correction process is driven by the most salient visual evidence, rather than dispersed and potentially irrelevant features.

\paragraph{Steering Distance for Inventation}
\begin{figure}
    \centering
    \includegraphics[width=1\linewidth]{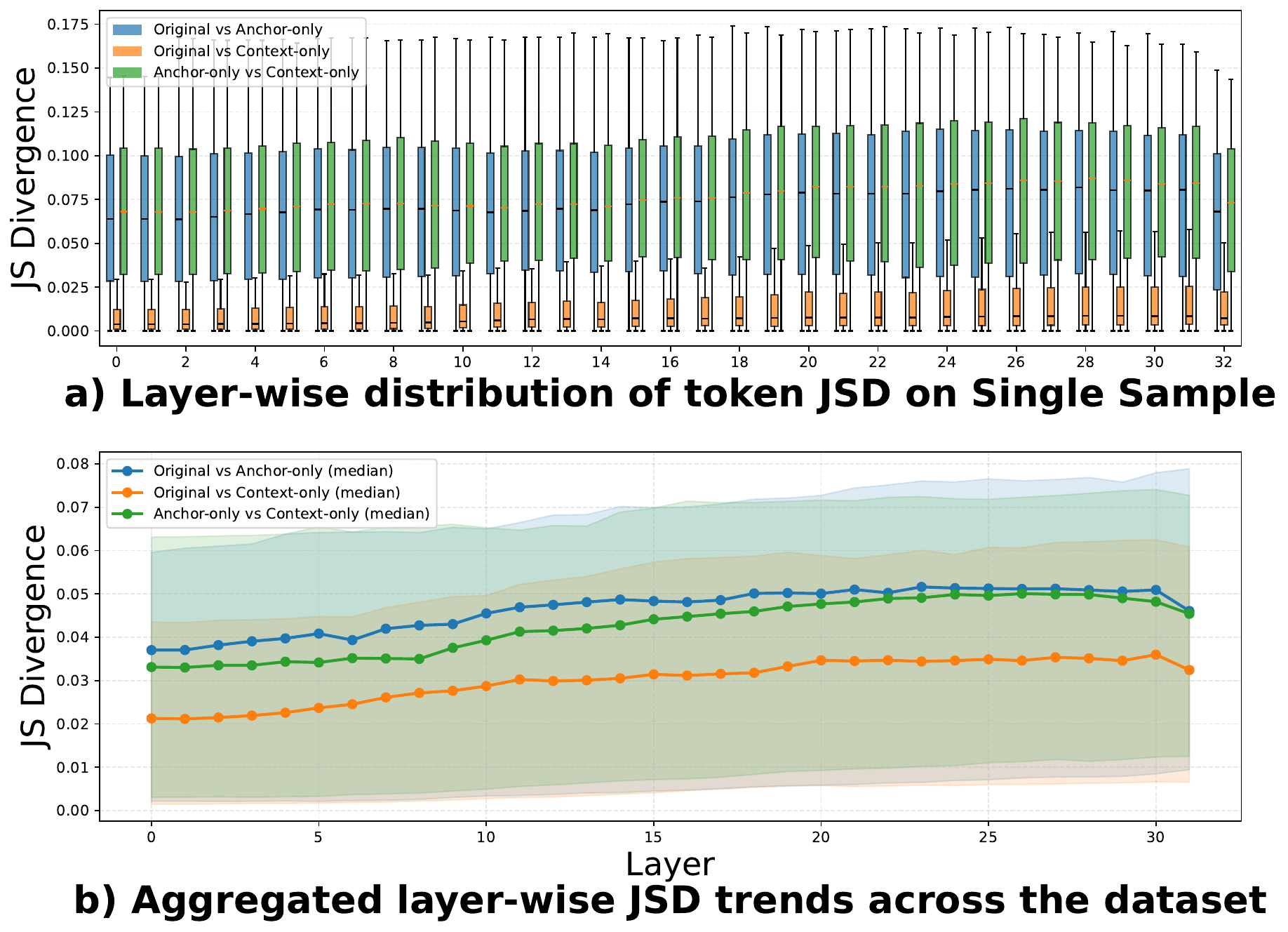}
    \caption{Layer-wise analysis of hidden state shifts.
(a) Presenting all tokens in a single representative sample from MMHal-Bench.
(b) Focusing on introspected tokens across the MMHal-Bench dataset.
The solid lines denote the median JS divergence, while the shaded regions indicate the interquartile range (IQR).}
    \label{fig:Layer-wise analysis of hidden state shifts}
\end{figure}
To analyze the reliability of our enhanced representations, we compute the token-wise JS divergence between the original hidden states and two counterfactual branches, Anchor-only and Context-only, across all decoder layers (Fig.~\ref{fig:Layer-wise analysis of hidden state shifts}).

In Fig.~\ref{fig:Layer-wise analysis of hidden state shifts}(a), the divergence between Original and Anchor-only states consistently exceeds that of the Context-only branch. This implies that removing the anchor alters representations significantly more than removing the context, suggesting the vanilla model is overly shaped by background patterns rather than the critical visual signal—a mechanism consistent with context-driven hallucination.
Fig.~\ref{fig:Layer-wise analysis of hidden state shifts}(b) confirms this trend at the dataset level. The median JS divergence for the Anchor-only path increases with depth, significantly surpassing the flat Context-only curve, which indicates an accumulation of context-driven bias. VLI counteracts this drift by explicitly steering hidden states toward the anchor branch, restoring visual grounding.
The minimal divergence between original and Context-only states suggests that task-irrelevant background often dominates the global representation. Conversely, the rising deviation of Anchor-only states reveals how accumulated background features progressively displace target semantics. Although this gap narrows in the final layers, the persistent misalignment highlights the necessity of bi-causal steering to reinforce the visual anchor against noise.
More analysis for steering JS divergence on all tokens across dataset can be seen in Appendix \ref{app:logit_diver}.

\subsection{Impact of Hyperparameters}
\label{sec:hyperparam}
\begin{figure}
    \centering
    \includegraphics[width=0.98\linewidth]{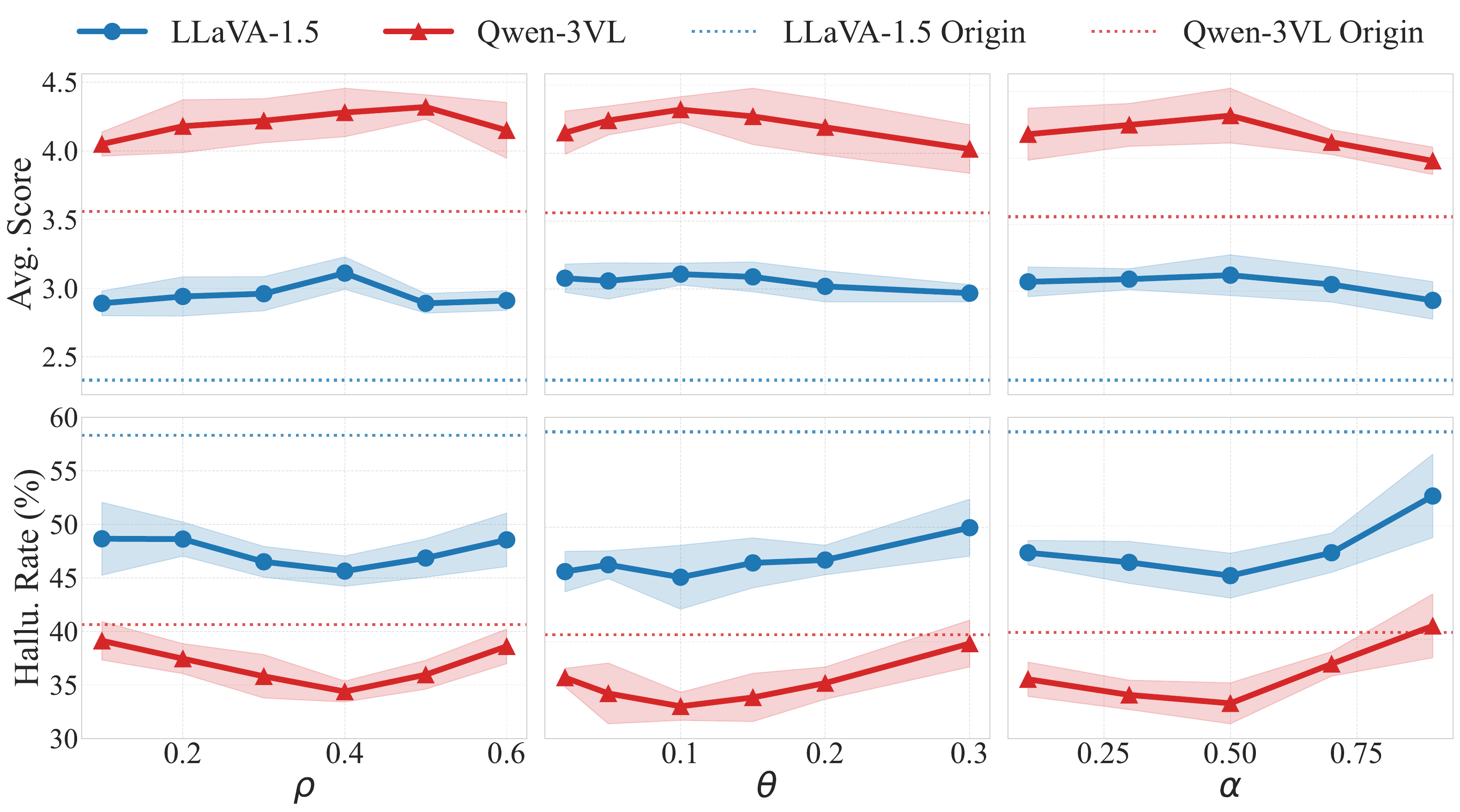}
    \caption{Hyperparameter sensitivity analysis on LLaVA-1.5 and Qwen3-VL. The solid lines represent the average performance of VLI, while the shaded regions indicate the standard deviation, highlighting the stability of our method. The dashed lines denote the baseline performance of the original models (Origin) without intervention.}
    \label{fig:hyper_param}
\end{figure}
We conduct a sensitivity analysis on three key hyperparameters: the cumulative energy ratio $\rho$ for \textit{Attributive Introspection}, the conflict risk threshold $\theta$ for triggering intervention, and the steering strength $\alpha$. The results, illustrated in Fig~\ref{fig:hyper_param}, demonstrate that VLI consistently outperforms the original model across a wide range of settings.

\noindent\textbf{Impact of $\rho$.} The parameter $\rho$ determines the spatial extent of the introspected anchor mask. Performance generally peaks at $\rho=0.4$, where LLaVA achieves a Score of 3.11 and a hallucination rate of 45.63\%. The narrow error bands around the curve suggest that the method remains robust to minor variations in anchor selection. Although Qwen3-VL achieves a slightly higher score at $\rho=0.5$, its hallucination rate remains lowest at $\rho=0.4$, confirming the robustness of this setting.

\noindent\textbf{Impact of $\theta$.} The risk threshold $\theta$ controls the sensitivity of the \textit{Attributive Introspection} phase. Both models achieve optimal performance at $\theta=0.10$, enabling Qwen3-VL to reach a peak Score of 4.32. Notably, the performance curve remains well above the baseline even at suboptimal thresholds, validating the efficacy of our intervention. However, increasing the threshold beyond 0.15 makes the model too conservative; at $\theta=0.30$, the LLaVA score drops to 2.97 while the hallucination rate rises to 49.95\%, narrowing the performance gap with the baseline.

\noindent\textbf{Impact of $\alpha$.} The steering strength $\alpha$ regulates the magnitude of the latent intervention. The results indicate that a strength of $\alpha=0.5$ yields the best balance, resulting in a Score of 3.12 for LLaVA and 4.32 for Qwen3-VL. This setting effectively corrects cognitive bias with high stability, as evidenced by the compact error bands. Conversely, excessive steering with $\alpha \ge 0.7$ harms model performance. Specifically, at $\alpha=0.9$, the LLaVA hallucination rate spikes to 52.65\%, causing the performance trajectory to sharply decline towards the baseline level, likely due to over-modification of the hidden states disrupting linguistic fluency.

\section{Conclusion}

In this paper, we introduced Vision-Language Introspection, a training-free framework designed to mitigate object hallucination by simulating metacognitive self-correction. VLI addresses the disconnect between linguistic priors and visual evidence by synergizing Attributive Introspection for causal anchor localization with Interpretable Bi-Causal Steering for latent representation rectification. This approach effectively isolates visual truths from background noise and neutralizes blind confidence without requiring parameter updates. Experimental results confirm that VLI achieves state-of-the-art performance on both discriminative and generative benchmarks, demonstrating that equipping multimodal models with introspective capabilities offers a robust pathway toward enhanced trustworthiness.

\section*{Limitations}

First, as analyzed in Appendix \ref{app:latency}, the construction of counterfactual states during the introspection process introduces additional computational overhead compared to standard decoding strategies. Although we implemented a parallel processing mechanism to mitigate this latency issue, this solution necessitates a significant increase in GPU memory consumption which may constrain deployment on resource-limited devices. Second, the effectiveness of our Attributive Introspection relies on the premise that the base model possesses identifiable expert attention heads that correctly align semantic concepts with visual regions. In scenarios involving highly abstract concepts or where the underlying model fails to form concentrated attention patterns, the precision of causal anchor extraction may degrade and consequently limit the efficacy of the steering intervention.





\bibliography{anthology,acl2021}
\bibliographystyle{acl_natbib}

\newpage
\appendix
\newpage 
\clearpage

\section{Theoretical Analysis}
\label{app:theory}

In this section, we provide a theoretical foundation for the VLI framework. We demonstrate that our Interpretable Bi-Causal Steering mechanism mathematically functions as a semantic contrastive filter that enhances the Signal-to-Noise Ratio (SNR) of visual representations while decoupling linguistic priors. Furthermore, we justify the Adaptive Confidence Calibration as a regularization of causal sensitivity.

\subsection{Latent Linear Representation Hypothesis}
Building on recent findings in mechanistic interpretability~\cite{park2025halloc, jiang2025devils}, we posit the \textit{Linear Representation Hypothesis}. We assume that at layer $l$, the high-dimensional latent state $h \in \mathbb{R}^d$ can be approximated as a linear superposition of independent semantic subspaces.
For a multimodal input $(V, T)$, we decompose the latent state $h$ into three orthogonal components:
\begin{equation}
    h \approx \mathbf{z}_{obj} + \mathbf{z}_{ctx} + \mathbf{z}_{lang}
\end{equation}
where:
\begin{itemize}
    \item $\mathbf{z}_{obj}$: The causal visual vector corresponding to the specific object features defined by the anchor mask $\mathcal{M}_s$.
    \item $\mathbf{z}_{ctx}$: The visual context vector (background noise/sinks) corresponding to $1-\mathcal{M}_s$.
    \item $\mathbf{z}_{lang}$: The linguistic vector encoding syntax and textual priors derived from $T_{<t}$.
\end{itemize}

Under the hallucination scenario, the model generation $y_t$ is dominated by $\mathbf{z}_{lang}$ and $\mathbf{z}_{ctx}$ (e.g., object co-occurrence priors in the background), while the grounded evidence $\mathbf{z}_{obj}$ is suppressed. Our goal is to rectify the distribution $P(y|h)$ to maximize the mutual information $I(y; \mathbf{z}_{obj})$.

\subsection{Derivation of Bi-Causal Steering}
VLI constructs two counterfactual states via inpainting: the Context-Only state $h_c$ (where the object is masked) and the Anchor-Only state $h_a$ (where the background is masked).

Assuming the inpainting operation $\mathcal{I}$ effectively suppresses the masked signal to a null vector $\mathbf{0}$ (or a mean vector orthogonal to the specific features), we can formulate these states as:
\begin{align}
    h_g &= \mathbf{z}_{obj} + \mathbf{z}_{ctx} + \mathbf{z}_{lang} \label{eq:theo_hg} \\
    h_c &\approx \mathbf{0} + \mathbf{z}_{ctx} + \mathbf{z}_{lang} \label{eq:theo_hc} \\
    h_a &\approx \mathbf{z}_{obj} + \mathbf{0} + \mathbf{z}_{lang} \label{eq:theo_ha}
\end{align}
Note that both counterfactual states retain the linguistic component $\mathbf{z}_{lang}$ because the textual input remains identical.

\paragraph{Decoupling Linguistic Priors.}
We define the steering vector $\Delta = h_a - h_c$. substituting Eq. \ref{eq:theo_hc} and \ref{eq:theo_ha}:
\begin{equation}
\begin{aligned}
    \Delta &= (\mathbf{z}_{obj} + \mathbf{z}_{lang}) - (\mathbf{z}_{ctx} + \mathbf{z}_{lang}) \\
           &= \mathbf{z}_{obj} - \mathbf{z}_{ctx}
\end{aligned}
\end{equation}
\textbf{Proposition 1 (Linguistic Orthogonality):} The steering vector $\Delta$ is orthogonal to the linguistic prior $\mathbf{z}_{lang}$.
This derivation proves that $\Delta$ captures a pure \textit{visual contrast}, the direction pointing towards the object and away from the background, while mathematically canceling out the linguistic priors. This explains why VLI does not degrade language fluency: the steering occurs solely in the visual semantic subspace.

\paragraph{Signal-to-Noise Ratio Enhancement.}
The rectified state is defined as $h_d = h_g + \alpha \Delta$. Substituting the components:
\begin{equation}
\begin{aligned}
    h_d &= (\mathbf{z}_{obj} + \mathbf{z}_{ctx} + \mathbf{z}_{lang}) + \alpha(\mathbf{z}_{obj} - \mathbf{z}_{ctx}) \\
        &= (1 + \alpha)\mathbf{z}_{obj} + (1 - \alpha)\mathbf{z}_{ctx} + \mathbf{z}_{lang}
\end{aligned}
\end{equation}
We define the Signal-to-Noise Ratio (SNR) of the visual representation as the ratio of the object magnitude to the context/noise magnitude: $\text{SNR}(h) = \frac{\|\mathbf{z}_{obj}\|}{\|\mathbf{z}_{ctx}\|}$.
Comparing the SNR of the grounded state $h_g$ and the rectified state $h_d$:
\begin{equation}
    \text{SNR}(h_d) = \frac{1+\alpha}{1-\alpha} \cdot \frac{\|\mathbf{z}_{obj}\|}{\|\mathbf{z}_{ctx}\|} = \frac{1+\alpha}{1-\alpha} \cdot \text{SNR}(h_g)
\end{equation}
For any steering strength $0 < \alpha < 1$, the gain factor $\frac{1+\alpha}{1-\alpha} > 1$.
\textbf{Proposition 2 (SNR Amplification):} The Bi-Causal Steering strictly increases the SNR of the latent state, forcing the model to attend to the causal anchor $\mathbf{z}_{obj}$ while suppressing the confounder $\mathbf{z}_{ctx}$.

\subsection{Theoretical Justification for Calibration}
The Adaptive Confidence Calibration (Eq. \ref{eq:T_calib}) scales the temperature based on the ratio of global conflict to local causal conflict. We formalize this ratio as the \textit{Ungrounded Certainty Ratio}.

Let $\mathcal{D}_{KL}(P || Q)$ denote the divergence. The numerator $\mathcal{C}(P_g, P_u)$ approximates the \textbf{Total Perceptual Sensitivity}, how much the model's belief changes given \textit{any} visual input versus no vision. The denominator $\mathcal{C}(P_a, P_c)$ approximates the \textbf{Causal Sensitivity}, how much the belief changes specifically due to the presence of the object anchor versus the background.
\begin{equation}
    R_{risk} = \frac{\text{Total Sensitivity}}{\text{Causal Sensitivity}} \approx \frac{\|\partial P / \partial V\|}{\|\partial P / \partial \mathbf{z}_{obj}\|}
\end{equation}
\textbf{Case Analysis:}
\begin{itemize}
    \item \textbf{Valid Recognition:} If the model truly sees the object, $\mathcal{C}(P_a, P_c)$ is high (strong causal link). The ratio $R_{risk}$ is low, resulting in $T_c \approx 1$. The distribution remains sharp.
    \item \textbf{Hallucination (Blind Confidence):} If the model predicts an object due to priors or background context, $\mathcal{C}(P_g, P_u)$ may be high (vision changes the prior), but $\mathcal{C}(P_a, P_c) \to 0$ (the specific object pixels do not drive the decision). Here, $R_{risk} \to \infty$.
\end{itemize}
Consequently, Eq. \ref{eq:T_calib} drives $T_c \gg 1$, maximizing the entropy of the output distribution $P_{corr}$. This theoretically proves that our calibration mechanism functions as a dynamic regularizer that penalizes predictions unsupported by specific, pixel-wise causal evidence.

\section{Introspection conflict Analysis}
\begin{figure*}[h!]
    \centering
    \includegraphics[width=1\linewidth]{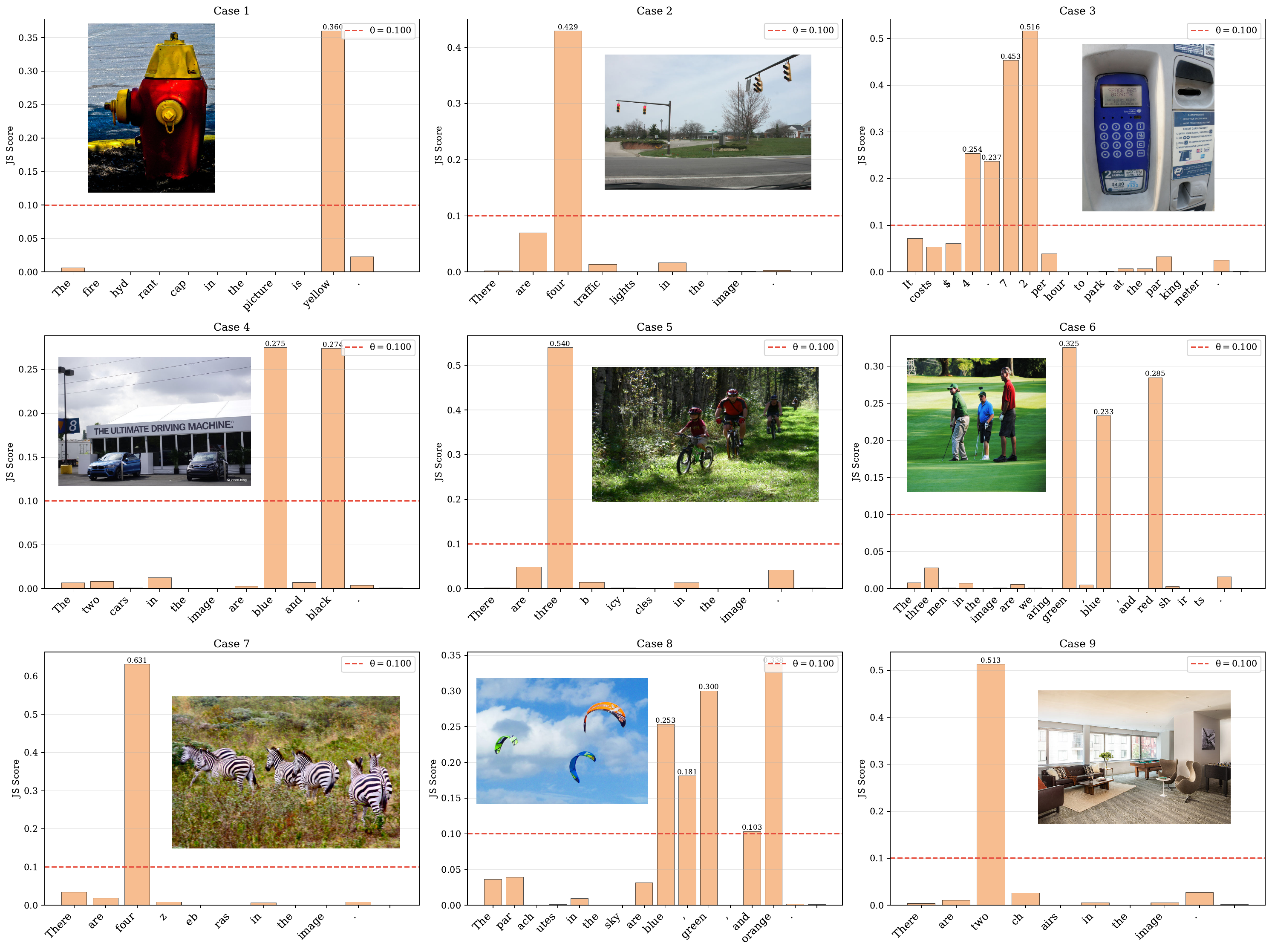}
    \caption{Introspective conflict scores between grounded and ungrounded decoding paths.
Token-wise introspective conflict scores of LLaVA-1.5 on nine representative MMHal-Bench samples. Each panel shows the input image together with the grounded answer, where the height of each bar indicates the JS divergence of the corresponding token between the grounded and ungrounded paths. The red dashed line marks the conflict-risk threshold $\theta = 0.10$ used to trigger introspection.}
    \label{fig:Introspective conflict scores between grounded and ungrounded decoding paths}
\end{figure*}

To gain a more concrete understanding of how the introspective conflict score behaves, we analyze nine representative MMHal-Bench examples in Fig. \ref{fig:Introspective conflict scores between grounded and ungrounded decoding paths}, each associated with a specific question. For each example, we compare the grounded decoding path, which conditions on the image, with the ungrounded path, which relies more heavily on language priors. The token-wise JS divergence between these two paths reflects how strongly the model’s belief changes once visual evidence is taken into account.

For Case 1 (What color is the fire hydrant cap in the picture?), the question explicitly asks for a color attribute. The conflict curve remains low for function words such as what, color, the, and in, but exhibits a sharp spike on the color token used in the grounded answer (e.g., yellow). This indicates that, without visual grounding, the model tends to follow a strong prior that fire hydrants are typically red, whereas the grounded path adjusts the answer to the correct but less frequent color, producing a high JS divergence exactly at the answer-bearing token. A similar pattern appears in Case 4 (What color are the two cars from right to left in the image?) and Case 6 (What are the colors of the shirts worn by the three men from left to right in the image?), where the largest conflicts occur on the specific color words describing each car or shirt. In Case 8 (What are the colors of the parachutes in the sky?), multiple color tokens show elevated divergence, reflecting that the grounded path must reconcile several distinct colors with the ungrounded prior that tends to favor a small set of frequent colors.

Counting questions show an analogous but complementary behavior. In Case 2 (How many traffic lights are there in the image?), the conflict scores for most tokens are near zero, but the numeral that encodes the predicted count (e.g., four) exhibits a prominent peak. This suggests that language priors alone do not confidently determine the number of traffic lights, and the grounded path must significantly adjust the count based on visual evidence. The same phenomenon is observed in Case 5 (How many bicycles are there in the image?), Case 7 (How many zebras are there in the image?), and Case 9 (How many chairs are there in the image?), where the highest JS divergence consistently concentrates on the numeral token that directly answers the question, while surrounding context tokens remain stable.

Case 3 (How much is it per hour to park at the parking meter?) further illustrates this behavior for fine-grained numeric attributes. Here, the model must output a specific price rather than a small integer count. The conflict curve stays low for the framing tokens (how much, per hour, to park), but spikes on the digits that compose the grounded hourly rate. This indicates that the visual reading of the meter substantially revises the ungrounded guess about the price, again localizing conflict to the answer-bearing portion of the sequence.

Across all nine cases, we observe a consistent sparsity pattern: most tokens have JS divergence well below the threshold 
$\theta=0.10$ , and only a small number of semantically critical tokens, colors, numerals, or digits that directly respond to the question, exceed this cutoff. This supports our design of the introspective conflict score as a selective trigger rather than a global perturbation: it remains quiet on benign context tokens and becomes active only where vision–language mismatch is likely to cause hallucination. Combined with our earlier hyperparameter study, which shows that thresholds in the vicinity of $\theta=0.1$ yield the best trade-off between hallucination reduction and overall performance, these case studies provide qualitative evidence that the chosen threshold is both effective and reasonable. It allows VLI to focus introspective interventions precisely on the answer tokens where correcting hallucinations matters most.

\begin{table*}[t]
\centering
\resizebox{\textwidth}{!}{%
\begin{tabular}{l|cc|cc|cc|cc}
\toprule
\multirow{2}{*}{\textbf{Method Configuration}} & \multicolumn{2}{c|}{\textbf{MMHal-Bench}} & \multicolumn{2}{c|}{\textbf{POPE (MSCOCO)}} & \multicolumn{2}{c|}{\textbf{POPE (A-OKVQA)}} & \multicolumn{2}{c}{\textbf{POPE (GQA)}} \\
 & Score $\uparrow$ & Hallu. Rate $\downarrow$ & Acc (\%) $\uparrow$ & F1 (\%) $\uparrow$ & Acc (\%) $\uparrow$ & F1 (\%) $\uparrow$ & Acc (\%) $\uparrow$ & F1 (\%) $\uparrow$ \\ \midrule
\multicolumn{9}{c}{\textit{Base Model: LLaVA-1.5}} \\ \midrule
Origin (Baseline) & 2.33 & 58.30 & 83.82 & 84.18 & 79.81 & 79.54 & 77.58 & 77.26 \\
\rowcolor{gray!10} \textbf{VLI (Standard)} & \textbf{3.11} & \textbf{45.63} & \textbf{89.27} & \textbf{89.61} & \textbf{85.87} & \textbf{85.54} & \textbf{83.18} & \textbf{83.49} \\
VLI + Explicit Sink Masking & 3.12 & 45.58 & 89.29 & 89.65 & 85.91 & 85.60 & 83.22 & 83.55 \\
VLI w/o Expert Heads (Avg) & 2.65 & 52.14 & 86.05 & 86.44 & 82.10 & 81.85 & 80.05 & 79.90 \\
VLI w/o Adaptive Threshold (Fixed-$k$) & 2.89 & 48.75 & 87.55 & 87.90 & 83.45 & 83.10 & 81.30 & 81.15 \\ \midrule
\multicolumn{9}{c}{\textit{Base Model: Qwen3-VL}} \\ \midrule
Origin (Baseline) & 3.56 & 40.63 & 90.53 & 91.14 & 86.82 & 87.13 & 82.34 & 81.93 \\
\rowcolor{gray!10} \textbf{VLI (Standard)} & \textbf{4.32} & \textbf{34.38} & \textbf{92.58} & \textbf{92.19} & \textbf{88.79} & \textbf{89.23} & \textbf{85.96} & \textbf{86.47} \\
VLI + Explicit Sink Masking & 4.33 & 34.32 & 92.61 & 92.25 & 88.85 & 89.30 & 86.01 & 86.52 \\
VLI w/o Expert Heads (Avg) & 3.88 & 37.95 & 91.20 & 91.55 & 87.40 & 87.65 & 83.50 & 83.80 \\
VLI w/o Adaptive Threshold (Fixed-$k$) & 4.10 & 36.12 & 91.85 & 91.80 & 88.05 & 88.45 & 84.80 & 85.10 \\ \bottomrule
\end{tabular}%
}
\caption{Robustness analysis against Visual Attention Sinks on MMHal-Bench and POPE. We compare our standard VLI against variants with explicit sink masking and ablated attention mechanisms. \textbf{VLI + Sink Masking} explicitly filters tokens based on Eq. \ref{eq:sink_way}. The negligible performance gap validates that VLI is intrinsically robust to attention sinks.}
\label{tab:sink_robustness}
\end{table*}

\section{Robustness to Visual Attention Sinks}
\label{app:sink}

Recent studies~\cite{kang2025see} identify the visual attention sink phenomenon in LVLMs, where specific irrelevant tokens such as delimiters or background patches disproportionately absorb attention mass. A standard identification method relies on detecting anomalous activation magnitudes:
\begin{equation}
\label{eq:sink_way}
    \phi(v_j) = \max_{d \in \mathcal{D}_{sink}} \frac{|v_j[d]|}{\|v_j\|_2}
\end{equation}
where tokens exceeding a threshold $\tau_{sink}$ are flagged as sinks.

While explicit masking of these sinks is a common remedy, our \textbf{Expert Head Selection} and \textbf{Cumulative Energy Thresholding} mechanisms described in Sec \ref{sec:anchor_extraction} provide intrinsic robustness against this noise without requiring a separate sink detection module. Since visual sinks typically do not align with the semantic regions required for grounded prediction, they are naturally filtered out during the expert head calibration phase characterized by low $\mu_{l,h}$ scores. Furthermore, our adaptive anchor extraction focuses on the cumulative probability mass $\rho$. Consequently, unless a sink dominates the global attention distribution to an extreme degree, which is a rare occurrence in the identified expert heads, it is excluded from the causal mask $\mathcal{M}_s$. Therefore, VLI efficiently purifies visual evidence without the computational overhead of explicit sink modeling.

To validate this, we compare our standard VLI framework against a variant augmented with explicit sink masking, denoted as \textit{VLI + Explicit Sink Masking}, and two ablated variants lacking our core filtering mechanisms. The results are reported in Table \ref{tab:sink_robustness}.

\noindent\textbf{Negligible Gain from Explicit Masking.}
Comparing \textit{VLI (Standard)} with \textit{VLI + Explicit Sink Masking}, we observe minimal performance differences across all metrics. For instance, on LLaVA-1.5, the hallucination rate on MMHal-Bench improves marginally from 45.63\% to 45.58\% where $\Delta < 0.1\%$, and POPE accuracy remains statistically stagnant. This confirms that the tokens identified as sinks by explicit algorithms are already being effectively filtered out by the internal mechanisms of VLI, rendering the additional computational overhead of sink detection redundant.

\noindent\textbf{Role of Expert Head Selection.}
The significant performance drop in \textit{VLI w/o Expert Heads}, exemplified by a 6.51\% increase in hallucination rate on LLaVA-1.5, highlights the critical role of our head selection strategy. Attention sinks typically manifest as high-magnitude activations across global average heads. By selectively aggregating attention only from \textit{expert heads} $\mathcal{H}_{expert}$ that demonstrate high localization accuracy $\mu_{l,h}$, VLI naturally bypasses heads dominated by sink tokens, thereby purifying the causal signal.

\noindent\textbf{Efficacy of Cumulative Energy Thresholding.}
Similarly, replacing our adaptive energy thresholding with a fixed top-$k$ strategy, referred to as \textit{VLI w/o Adaptive Threshold}, leads to a noticeable performance degradation. Fixed-$k$ selection risks including high-activation sink tokens that may appear in the long tail of the distribution, or excluding valid semantic regions when the object is large. Our cumulative energy approach utilizing $\rho$ ensures that the anchor mask $\mathcal{M}_s$ locks onto the semantic core, naturally excluding sink tokens unless they dominate the probability mass. Such domination is a rarity within identified expert heads.

In conclusion, VLI achieves robustness to visual attention sinks not through external patching, but through the synergistic design of expert head selection and adaptive anchor extraction.

\begin{figure}[h!]
    \centering
    \includegraphics[width=1\linewidth]{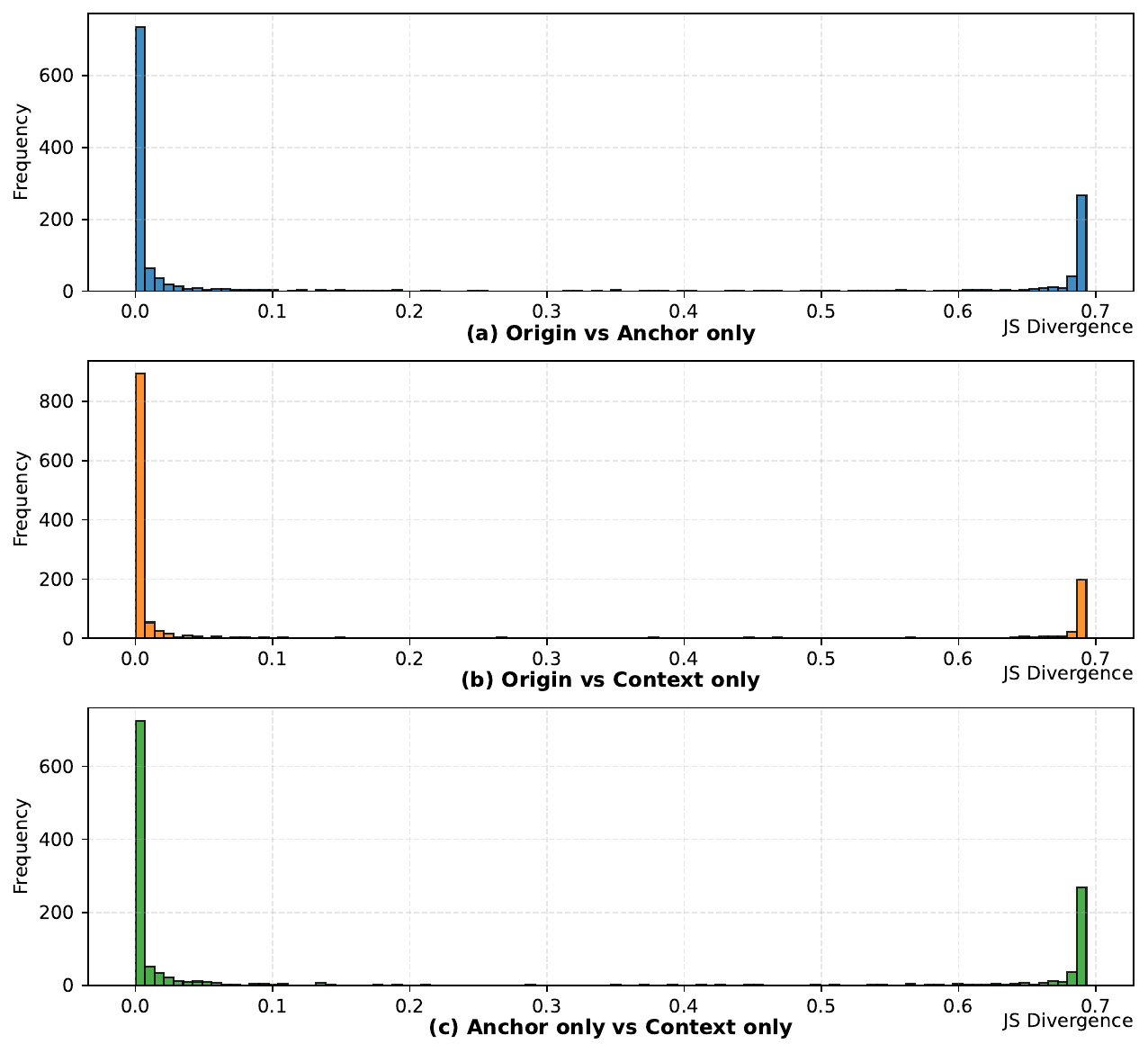}
    \caption{Distribution of Jensen–Shannon divergence between the logits produced by the original model and those produced using the enhanced input across all samples.}
    \label{fig:logits}
\end{figure}

\section{Logits Divergence Analysis}
\label{app:logit_diver}

Fig. ~\ref{fig:logits} presents the histograms of Jensen-Shannon (JS) divergence between the logits of the original decoding path and the counterfactual steering branches. The distribution exhibits a pronounced bimodality characterized by a heavy concentration of mass near zero and a distinct, sparse peak in the high-divergence region around 0.7. This statistical behavior provides empirical validation for the three core components of the VLI framework: Attributive Introspection, Bi-Causal Steering, and Adaptive Confidence Calibration.

First, the overwhelming density near zero divergence in Fig.~\ref{fig:logits}(a) and Fig.~\ref{fig:logits}(b) indicates that for the vast majority of generated tokens, the linguistic priors and visual context remain consistent. These tokens correspond to functional words or unambiguous context, as qualitatively illustrated in Fig.~6, where tokens such as \textit{there}, \textit{are}, and \textit{in} exhibit negligible conflict scores. This functional sparsity confirms that computationally expensive interventions are unnecessary for most generation steps, justifying the efficiency of our selective introspection mechanism which only triggers upon detecting conflict.

Second, the secondary peak in the high-divergence region signifies the existence of a specific subset of tokens where the visual anchor strictly contradicts the background context and linguistic priors. This aligns with the findings in Fig.~\ref{fig:Layer-wise analysis of hidden state shifts}(b), where the Anchor-only hidden states progressively deviate from the Context-only states across model layers. The sharp separation in Fig.~\ref{fig:logits}(c) between Anchor-only and Context-only logits proves that hallucinations are not caused by global degradation but by specific conflicts where the background noise overwhelms the object signal. This precise separation is critical for the efficacy of our Bi-Causal Steering vector $\Delta h$, ensuring that the intervention vector is orthogonal to the linguistic subspace and targets only the semantic misalignment.

Finally, the clear bimodality supports the design of the Adaptive Confidence Calibration mechanism. The discrete nature of the high-divergence mode suggests that hallucination is a binary state change rather than a linear degradation. Consequently, the use of the hyperbolic tangent function in Eq.~\ref{eq:T_calib} is theoretically sound, as it functions as a soft-gate that rapidly penalizes confidence only when the token falls into this high-risk distribution tail. The ablation study in Table~\ref{tab:ablation_study} further corroborates this; the removal of the Anchor-only component results in the most significant performance drop because it eliminates the high-divergence signal necessary to counteract the context-driven hallucinations prevalent in the original distribution.



\section{Latency and Computational Cost Analysis}
\label{app:latency}










\begin{table*}[htbp]
\centering
\resizebox{\textwidth}{!}{%
\begin{tabular}{l|cccccc|cc}
\toprule
\multirow{2}{*}{\textbf{Method}} & \multicolumn{6}{c|}{\textbf{Efficiency Metrics}} & \multicolumn{2}{c}{\textbf{Performance (MMHal)}} \\
\cmidrule(lr){2-7} \cmidrule(lr){8-9}
 & \textbf{Total} & \textbf{Total} & \textbf{Latency} & \textbf{Relative} & \textbf{Throughput} & \textbf{Memory} & \textbf{Hallucination} & \textbf{Overall} \\
 & \textbf{Runtime (s)} & \textbf{Tokens} & (ms/token) $\downarrow$ & \textbf{Cost} $\downarrow$ & (tokens/s) $\uparrow$ & \textbf{Overhead} & \textbf{Rate (\%)} $\downarrow$ & \textbf{Score} $\uparrow$ \\
\midrule
\rowcolor{gray!10} \textbf{Origin} (Baseline) & 3.130 & 70 & \textbf{44.71} & \textbf{1.00$\times$} & \textbf{22.37} & 1.0$\times$ & 58.30 & 2.33 \\
\midrule
VCD & 6.179 & 81 & 76.28 & 1.71$\times$ & 13.11 & $\sim$2.0$\times$ & 63.54 & 2.46 \\
CICD & 9.308 & 80 & 116.35 & 2.60$\times$ & 8.59 & $\sim$2.0$\times$ & 58.33 & 2.19 \\
ClearSight & 66.119 & 80 & 826.49 & 18.49$\times$ & 1.21 & $\sim$1.0$\times$ & 57.29 & 2.16 \\
OPERA & 30.011 & 74 & 405.56 & 9.07$\times$ & 2.47 & $\sim$1.0$\times$ & 58.30 & 2.40 \\
VTI & 6.853 & 80 & 85.66 & 1.92$\times$ & 11.67 & 1.0$\times$ & 51.00 & 2.39 \\
Nullu & 9.804 & 74 & 132.48 & 2.96$\times$ & 7.55 & 1.0$\times$ & 54.17 & 2.30 \\
\midrule
VLI (Serial) & 17.730 & 82 & 216.22 & 4.84$\times$ & 4.62 & 1.0$\times$ & \textbf{45.63} & \textbf{3.11} \\
\rowcolor{blue!5} \textbf{VLI (Parallel)} & 7.823 & 82 & \underline{95.41} & \underline{2.13$\times$} & \underline{10.48} & $\sim$3.0$\times$ & \textbf{45.63} & \textbf{3.11} \\
\bottomrule
\end{tabular}%
}
\caption{\textbf{Comprehensive Efficiency and Performance Analysis.} 
This table integrates raw runtime data with derived efficiency metrics and performance outcomes. 
\textit{Relative Cost} denotes the latency multiplier relative to the Origin model. 
\textit{Memory Overhead} is estimated based on the requirement for parallel decoding streams (e.g., VLI Parallel processes anchor, context, and grounded states simultaneously).}
\label{tab:latency_comparison}
\end{table*}

\begin{figure*}[h]
    \centering
    \begin{minipage}{0.6\linewidth}
        \vspace{3pt}
        \centerline{\includegraphics[width=\textwidth]{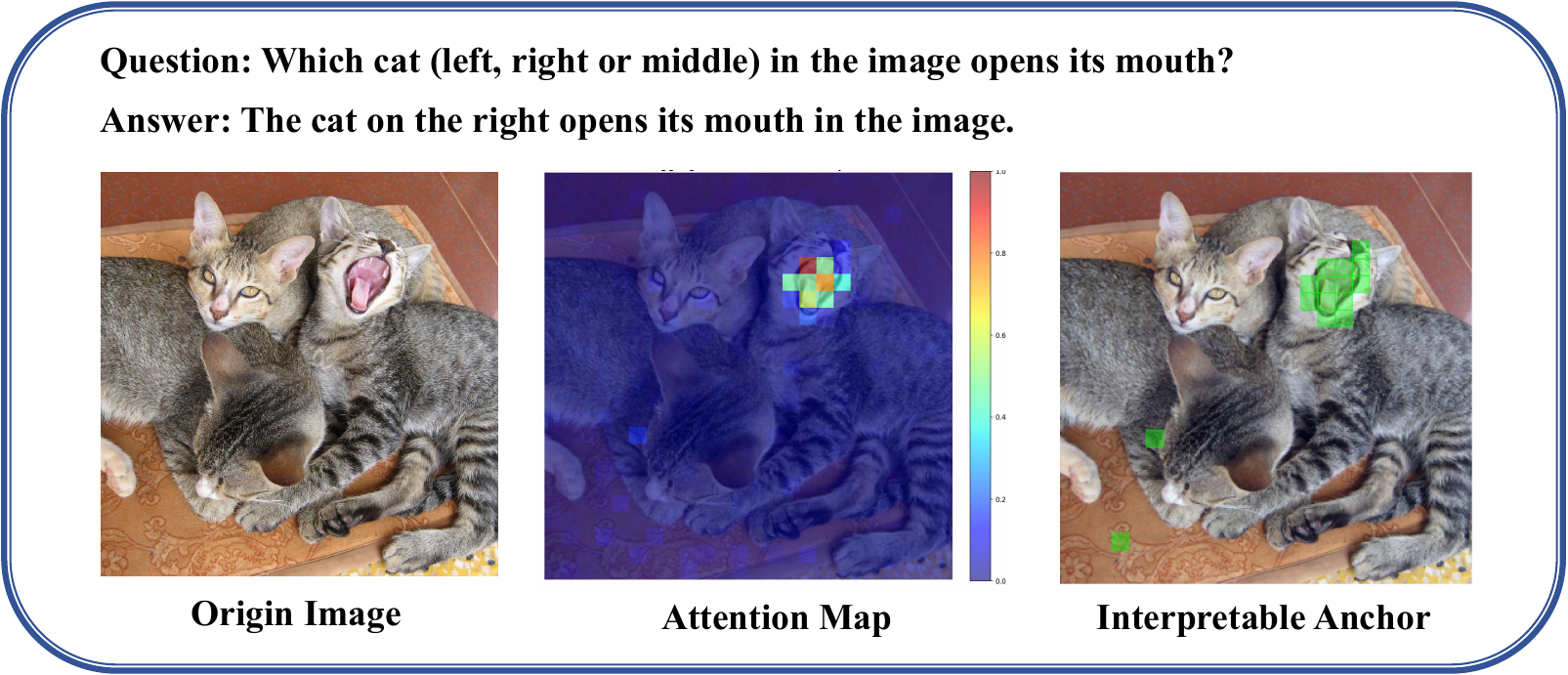}}

        \vspace{3pt}
        \centerline{\includegraphics[width=\textwidth]{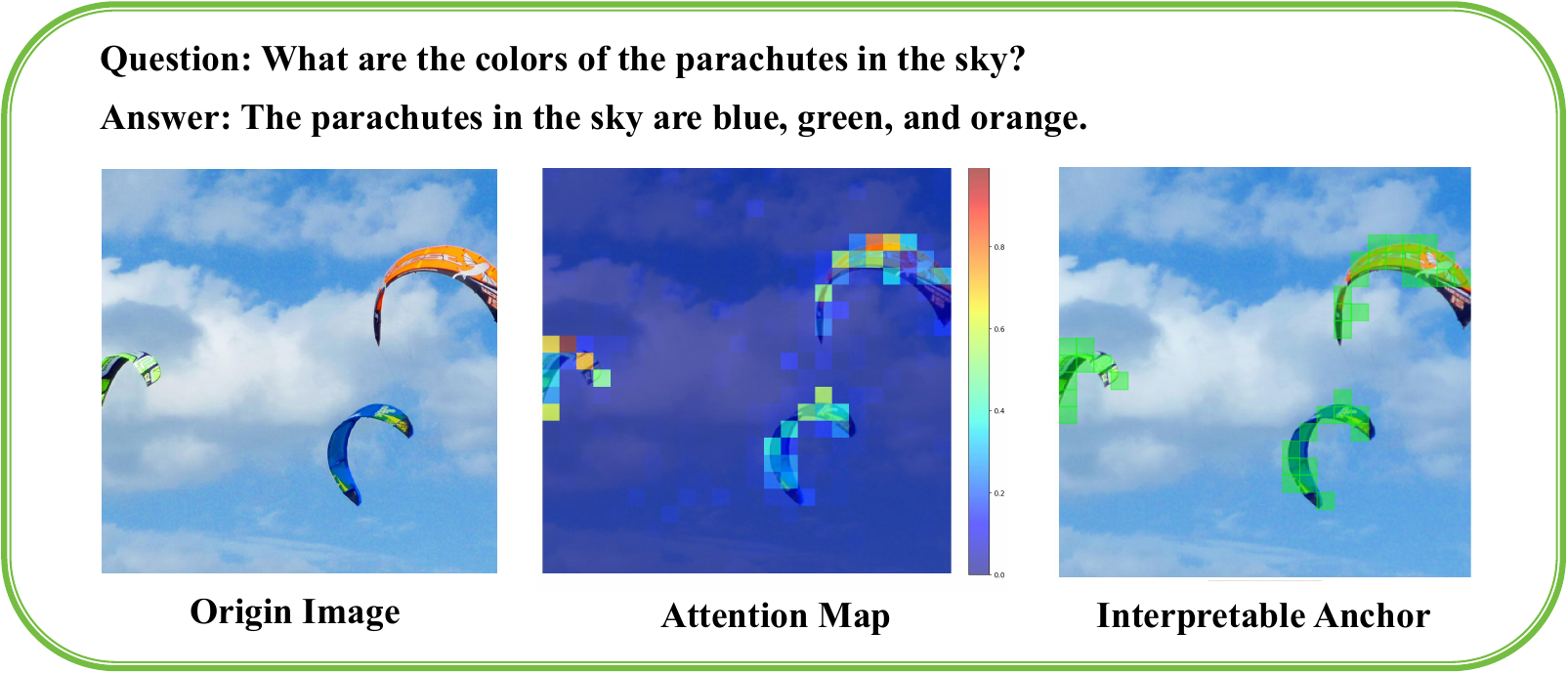}}

        \vspace{3pt}
        \centerline{\includegraphics[width=\textwidth]{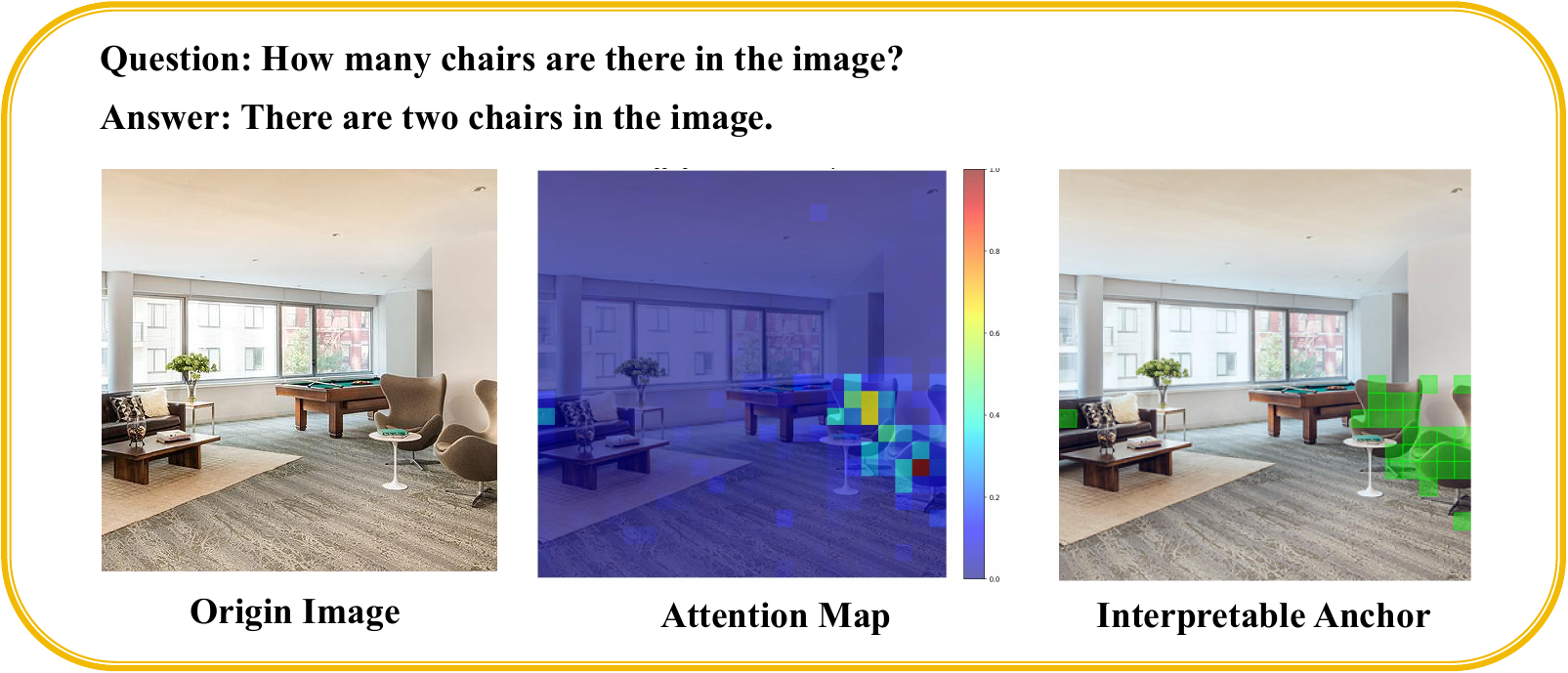}}

        \vspace{3pt}
        \centerline{\includegraphics[width=\textwidth]{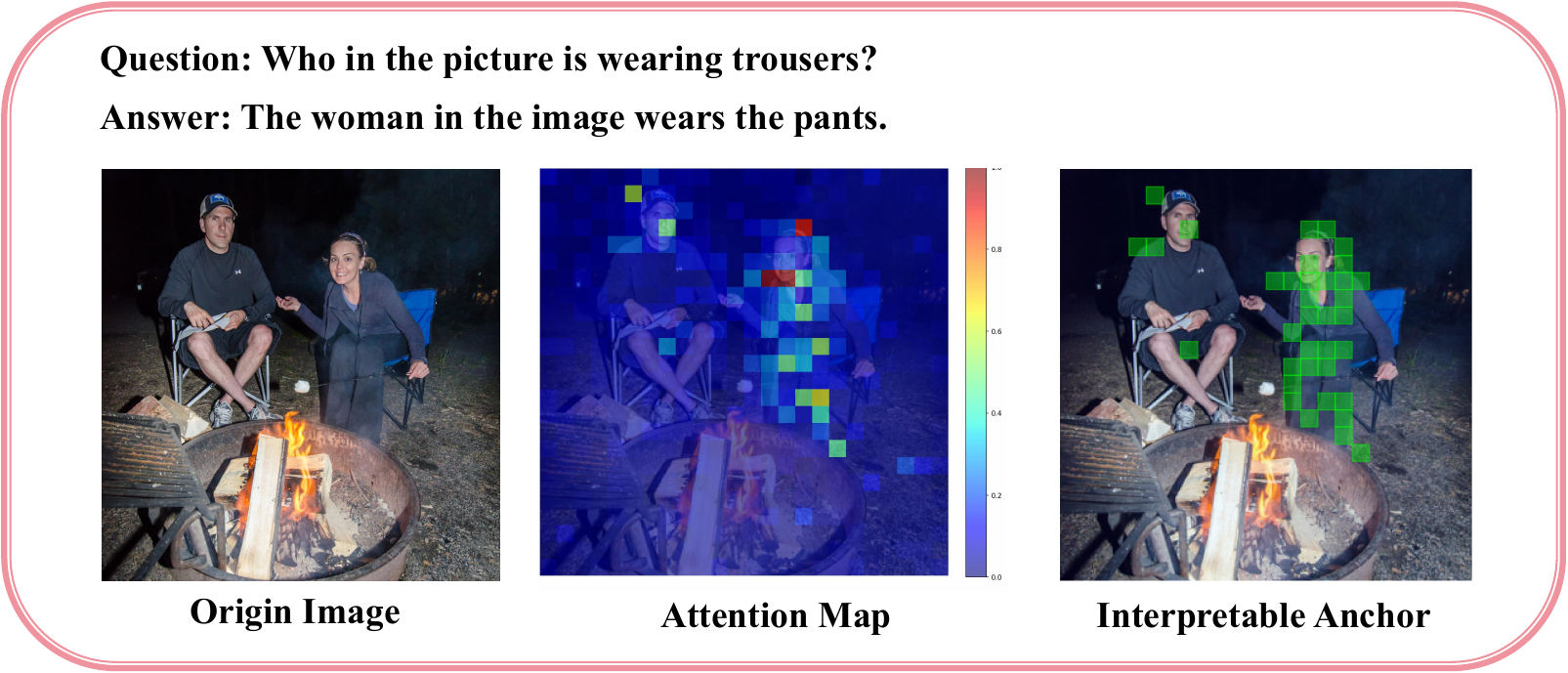}}

        \vspace{3pt}
        \centerline{\includegraphics[width=\textwidth]{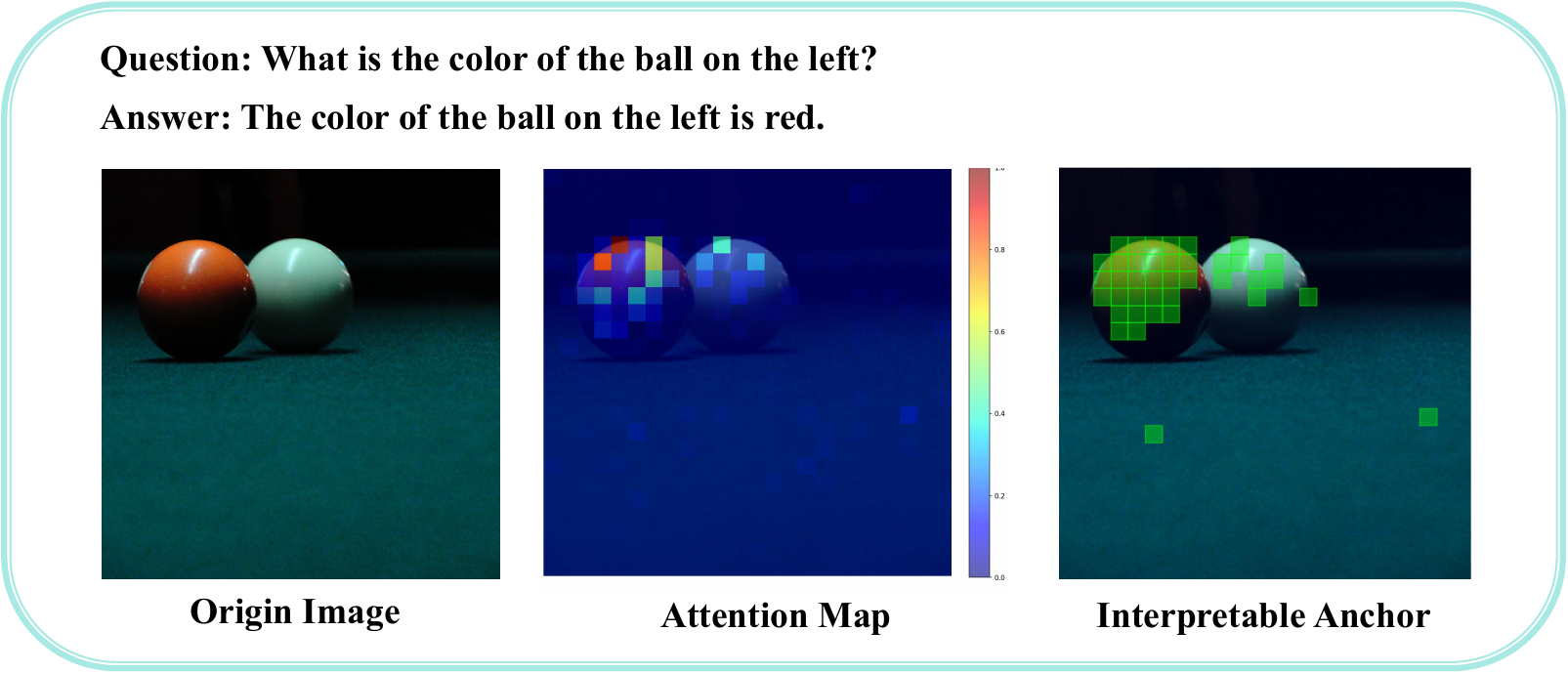}}
    \end{minipage}
    \caption{Visual comparisons of introspected visual anchor patterns under different scenarios.}
    \label{fig:introspect_case_vertical}
\end{figure*}

Table~\ref{tab:latency_comparison} presents a comprehensive latency comparison across different hallucination mitigation paradigms. Our proposed VLI framework, when optimized, achieves a competitive latency profile that balances computational efficiency with the depth of cognitive introspection.

\paragraph{Serial vs. Parallel Inference.} 
As indicated in Table~\ref{tab:latency_comparison}, the naive serial implementation of VLI (\textit{VLI Serial}) results in an average per-token latency of 216.22 ms. This increased latency is inherent to the \textit{Bi-Causal Steering} mechanism, which requires the computation of two additional counterfactual distinct states—Anchor-Only ($h_a$) and Context-Only ($h_c$)—alongside the original decoding path to derive the steering vector $\Delta h$. In a serial execution regime, these forward passes are computed sequentially, effectively tripling the inference cost for every generated token where introspection is triggered.

To mitigate this bottleneck, we implement a parallel processing mechanism (\textit{VLI Parallel}) that reduces the average per-token latency to 95.41 ms. This represents a speedup of approximately $2.27\times$ compared to the serial version, bringing our method's latency close to that of standard contrastive decoding methods like VCD (76.28 ms) and significantly lower than heavy attention-intervention methods such as OPERA (405.56 ms) or ClearSight (826.49 ms).

\paragraph{Parallelization Mechanism.} 
The parallelization is achieved by exploiting the independence of the counterfactual branches. During the \textit{Interpretable Bi-Causal Steering} phase, the computation of the anchor-driven state $h_a$ and the context-driven state $h_c$ does not depend on their mutual intermediate outcomes within the same step. Consequently, we construct a consolidated batch input $V_{batch} = [V_{original}; V_{anchor}; V_{context}]$ effectively performing the forward passes for all three representations simultaneously within a single GPU operation. This allows VLI to leverage the massive parallelism of modern hardware, masking the latency overhead of the additional forward passes.

\paragraph{Memory Overhead and Trade-offs.} 
While parallelization significantly reduces inference time, it introduces a trade-off regarding memory consumption. By processing the original and counterfactual streams in a single batch, the peak memory footprint increases during the steering phase. This overhead stems primarily from two sources:
\begin{enumerate}
    \item \textbf{Activation Storage:} The model must store intermediate activations for three concurrent streams instead of one, essentially tripling the memory required for temporary tensors during the forward pass.
    \item \textbf{KV-Cache Expansion:} To maintain context for the counterfactual paths across generation steps, the Key-Value (KV) cache must be maintained for the $h_a$ and $h_c$ branches in addition to the $h_g$ branch. This results in a linear increase in video memory usage proportional to the number of parallel streams.
\end{enumerate}
Despite this increased memory demand, the parallelized VLI framework remains deployable on standard academic hardware setups. The analysis confirms that by accepting a manageable increase in memory occupancy, VLI achieves a favorable sweet spot: it \textbf{delivers state-of-the-art hallucination reduction with a latency cost only marginally higher than simple contrastive baselines}, avoiding the prohibitive slowness of iterative attention-editing approaches.

\section{Case Study}\label{app:case_study}
\subsection{Interpretable Anchors in Attributive Introspection Phase}
\label{app:introspect_case_study}

In the \textit{Attributive Introspection} phase, the primary objective of VLI is to trace abstract cognitive dissonance, where linguistic priors conflict with sensory inputs, back to concrete regions in the input image. This process transforms latent cognitive uncertainty into explicit, interpretable visual evidence. This introspection culminates in the construction of a pixel-precise \textbf{Causal Anchor Mask}, which provides a grounded basis for the subsequent bi-causal steering.

The core intermediate representation in this phase is the Purified Attention Map, produced by the Attention Purification module. This module is designed to eliminate systemic and structural biases in raw attention by combining Expert Head Selection with Visual Sink Suppression. The resulting purified attention more faithfully reflects genuine visual grounding relevant to the token triggering the conflict.

Based on this purified signal, VLI generates the Causal Anchor Mask via a Cumulative Energy Thresholding strategy. Rather than relying on fixed thresholds, this adaptive mechanism selects the minimal set of pixels whose cumulative attention energy explains the dominant semantics of the introspection target, while effectively suppressing long-tail noise. The resulting binary mask thus captures only the visually critical regions 
responsible for verifying the model’s prediction.

As illustrated in Fig. \ref{fig:introspect_case_vertical}, VLI consistently localizes interpretable anchors across diverse visual reasoning tasks. In counting tasks (Case 3: determining the number of chairs), the framework introspects and highlights distinct object contours corresponding to each counted instance. For attribute recognition tasks (Case 2: identifying parachute colors and Case 5: identifying the color of the left ball), VLI isolates multiple spatially dispersed semantic targets across a wide field of view. In fine-grained detail recognition scenarios (Case 1: identifying which cat has its mouth open and Case 4: identifying who is wearing trousers), VLI precisely focuses on the relevant anatomical region rather than the entire object, demonstrating high-resolution semantic alignment.
These results validate that VLI moves beyond black-box failure analysis by establishing an explicit correlation between high-conflict tokens and their corresponding visual stimuli.

By explicitly mapping internal cognitive conflicts to external visual regions, VLI moves beyond treating hallucinations as black-box failures. The resulting anchor masks provide direct visual evidence that the identified tokens are intrinsically correlated with specific visual anchors in the input, validating the introspection process.

\subsection{Interpretable Bi-Causal Steering in Intervention Phase}
\label{app:suppress_case_study}

\begin{figure*}[t]
    \centering
    \includegraphics[width=1\textwidth]{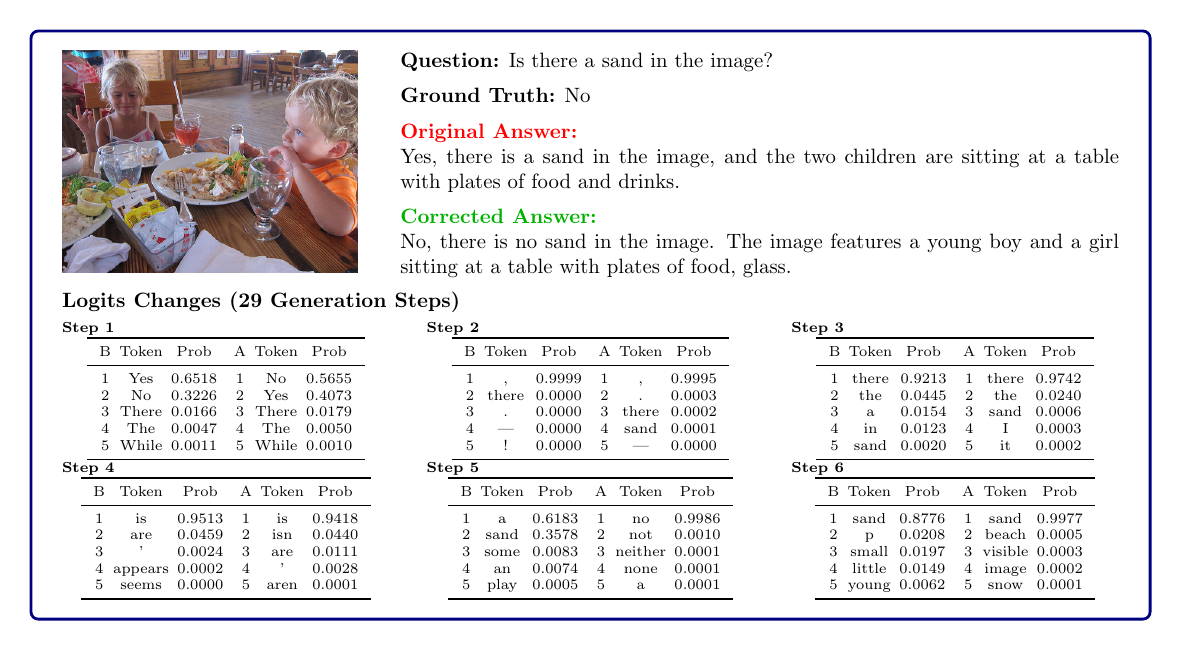}
    \caption{Case 1 from the POPE Random split (GQA subset).}
    \label{fig:exp1}
\end{figure*}

\begin{figure*}[t]
    \centering
    \includegraphics[width=1\textwidth]{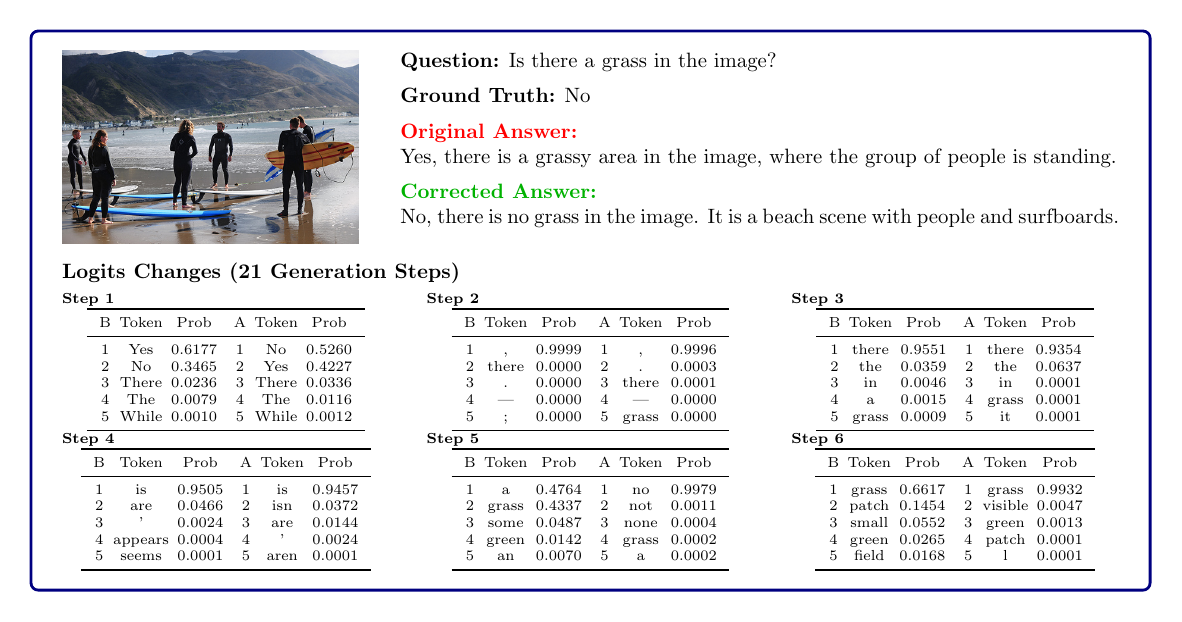}
    \caption{Case 2 from the POPE adversarial split (GQA subset).}
    \label{fig:exp2}
\end{figure*}

\begin{figure*}
    \centering
    \includegraphics[width=1\textwidth]{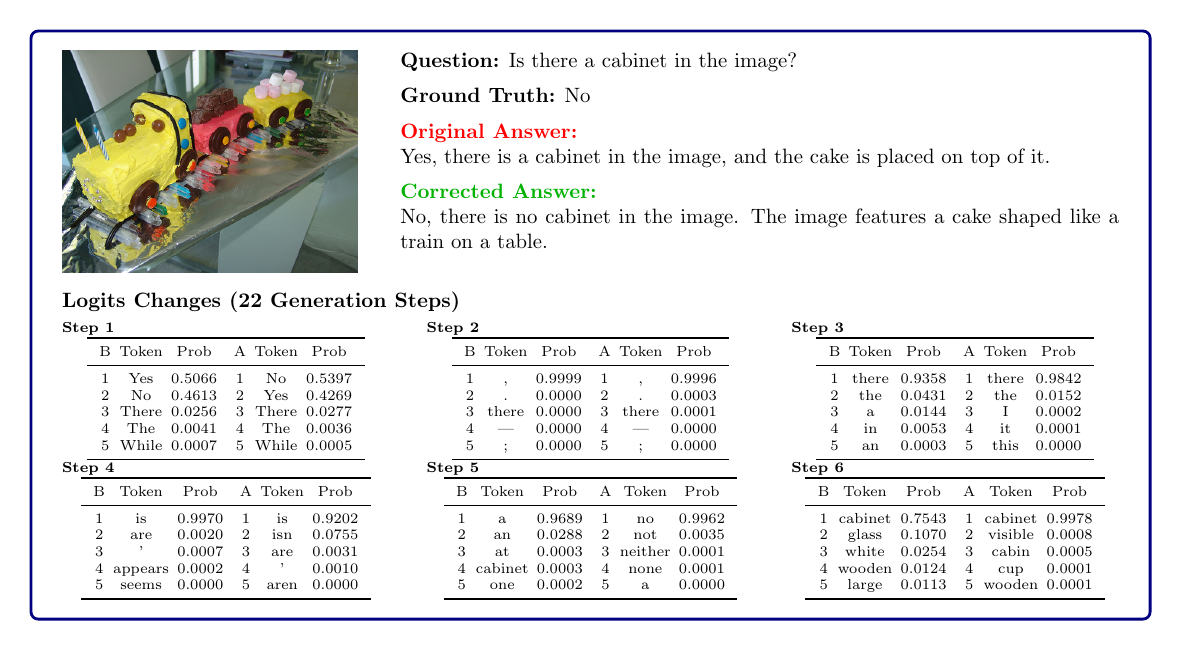}
    \caption{Case 3 from the POPE adversarial split (GQA subset).}
    \label{fig:exp3}
\end{figure*}

\begin{figure*}
    \centering
    \includegraphics[width=1\textwidth]{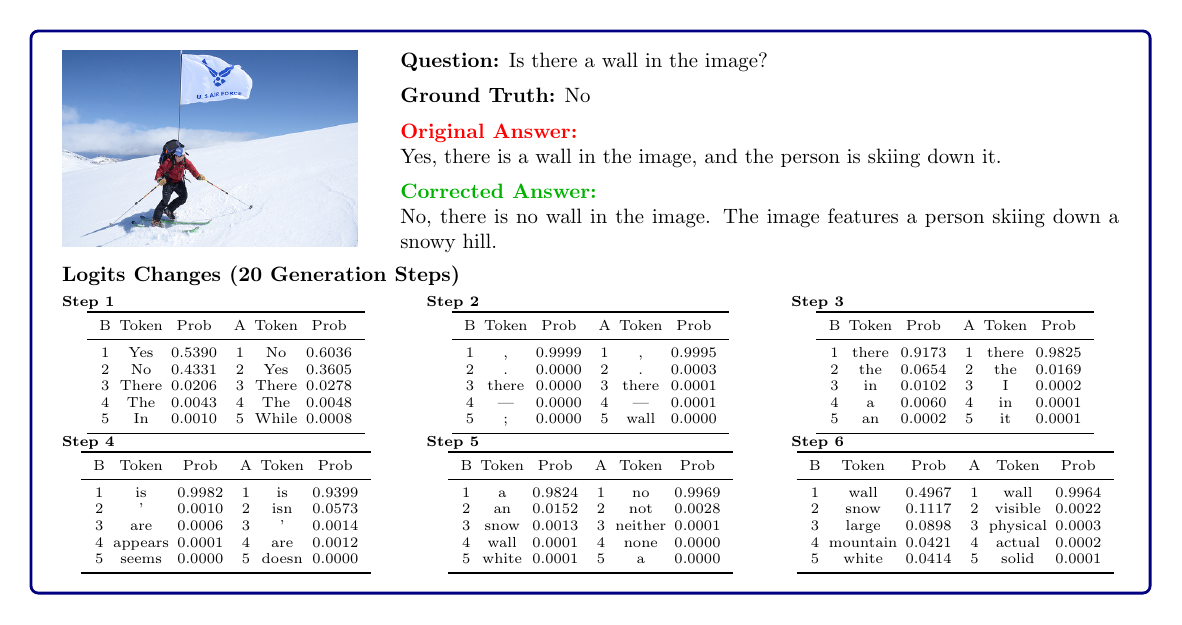}
    \caption{Case 4 from the POPE adversarial split (GQA subset).}
    \label{fig:exp4}
\end{figure*}

\begin{figure*}
    \centering
    \includegraphics[width=1\textwidth]{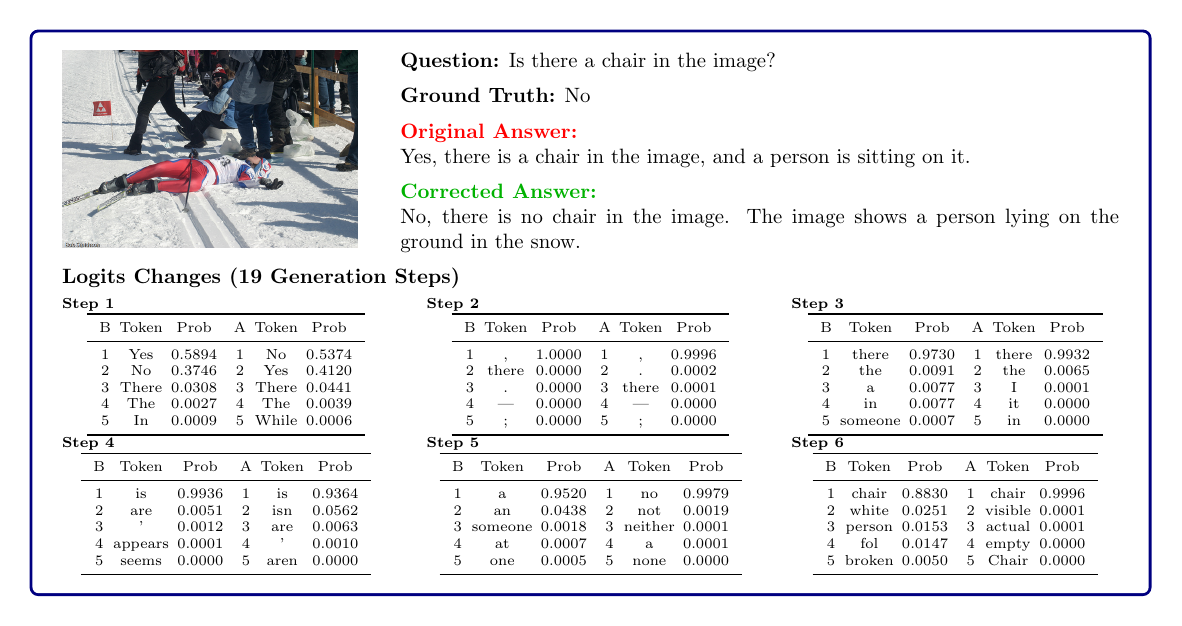}
    \caption{Case 5 from the POPE popular split (COCO subset).}
    \label{fig:exp5}
\end{figure*}

We provide additional examples through Fig.~\ref{fig:exp1} to Fig.~\ref{fig:exp5} showing the effect of the \textit{Interpretable Bi-Causal Steering} phase by detailing the evolution of logits during the generation process. These examples illustrate that VLI exerts a substantial influence on the final logit distribution by dynamically contrasting evidence against background noise. By effectively rectifying the probability bias at critical decision steps, our approach successfully steers the model away from overconfident linguistic priors and ensures the output aligns with the visual facts.

In Fig.~\ref{fig:exp1}, despite the clear dining setting, the baseline model hallucinates the presence of sand, initiating its response with "Yes" (0.6518). VLI effectively reverses this error at the logit level. At Step 1, it suppresses the "Yes" token to 0.4073 and elevates the correct "No" token to 0.5655. Furthermore, at Step 5, while the baseline model remains ambiguous, predicting "a" (0.6183) or "sand" (0.3578), our method solidly predicts "no" with a near-certain probability of 0.9986, ensuring the generated answer accurately states "there is no sand".

Fig.~\ref{fig:exp2} involves an image of surfers standing on a sandy beach, prompting the question, "Is there a grass in the image?". The baseline model misinterprets the texture of the ground, leading to an object hallucination where it asserts, "Yes, there is a grassy area". The logits table reveals that the baseline model strongly commits to this error at Step 1, assigning a probability of 0.6177 to the token "Yes". VLI successfully rectifies this at the onset by suppressing the affirmative response and elevating the correct token "No" to the top rank with a probability of 0.5260. The intervention remains robust at Step 5, where our method assigns a near-certain probability of 0.9979 to the token "no" (completing the phrase "there is no grass"), whereas the baseline remains confused, splitting its probability between "a" (0.4764) and "grass" (0.4337).

Visual ambiguity can often trigger hallucinations, as demonstrated by Fig.~\ref{fig:exp3} of a train-shaped cake, where the supporting table is mislabeled as a "cabinet". Unlike the previous clear-cut errors, the baseline here is initially ambivalent, only marginally preferring the incorrect "Yes" (0.5066) over "No" (0.4613) at Step 1. VLI proves decisive in these borderline cases. By effectively inverting the probability distribution via anchor reinforcement, it secures the correct trajectory with a "No" prediction (0.5397). This early intervention prevents the cascade of errors seen in the baseline, which at Step 5 firmly commits to the hallucination (predicting "a" with 0.9689 probability). In contrast, our method solidifies the correction with a definitive "no" (0.9962), ensuring the generated caption accurately reflects the scene.

Fig.~\ref{fig:exp4} illustrates a scenario where the model misinterprets the geometry of the scene, confusing a steep snowy slope for a vertical structure. When asked, "Is there a wall in the image?", the baseline model incorrectly affirms the presence of a wall, likely conflating the solid white expanse of the hill with a built barrier. The logit analysis at Step 1 shows the baseline favoring the hallucination with a "Yes" probability of 0.5390, compared to 0.4331 for "No". VLI effectively intervenes to correct this spatial misunderstanding. It shifts the probability distribution to favor "No" (0.6036), suppressing the "Yes" token to 0.3605. This correction prevents the model from constructing the erroneous phrase "a wall"; instead, at Step 5, our method assigns a decisive 0.9969 probability to the token "no", ensuring the final output correctly reflects the absence of the object.

Fig.~\ref{fig:exp5} presents a challenging case involving a skier lying on the snow, where the background contains wooden fencing and barriers that act as visual confounders. The baseline model is misled by these wooden textures, likely misidentifying the wooden slats as parts of a chair, and confidently predicts "Yes" (0.5894) at Step 1. VLI demonstrates its specific capability to target and resolve such ambiguity. By identifying the causal visual regions responsible for this confusion and enhancing the semantic contrast, VLI effectively suppresses the activation of erroneous features associated with the wooden elements. Consequently, the logit distribution at Step 1 is corrected to favor "No" (0.5374). This targeted intervention ensures that the model distinguishes the background noise from the foreground subject, resulting in a firm "no" prediction (0.9979) at Step 5 and a correct description of the person lying on the ground.


\section{Usage of AI Assistant}
The authors acknowledge the use of Gemini 3 Pro to assist with language editing and grammatical corrections. We affirm that the AI tool was not involved in the generation of scientific ideas, formulation of the methodology, or interpretation of the data. All intellectual content remains the work of the human authors.




\end{document}